\documentclass[lettersize,journal]{IEEEtran}
\usepackage{amsmath,amsfonts}
\usepackage{array}
\usepackage{textcomp}
\usepackage{stfloats}
\usepackage{url}
\usepackage{verbatim}
\usepackage{graphicx}
\usepackage{cite}
\usepackage{xcolor}

\usepackage{algorithmic}
\usepackage[ruled,vlined,linesnumbered]{algorithm2e}
\SetKwProg{Init}{init}{}{}

\newcommand\hl[1]{%
  \bgroup
  \hskip0pt\color{red!80!black}%
  #1%
  \egroup
}

\newcommand\hlb[1]{%
  \bgroup
  \hskip0pt\color{blue!80!black}%
  #1%
  \egroup
}

\usepackage{array}
\usepackage{longtable}
\usepackage{multirow}
\usepackage{tabularray}
\usepackage{subcaption}

\begin{document}

\title{
Overcoming Data Limitations in Internet Traffic Forecasting: LSTM Models with Transfer Learning and Wavelet Augmentation
}

\author{Sajal Saha, Anwar Haque, and Greg Sidebottom
		\thanks{S. Saha is with the Department of Computer Science, University of Northern British Columbia, Prince George, BC, V2N 4Z9, Canada. e-mail: sajal.saha@unbc.ca}

        \thanks{A. Haque is with the Department of Computer Science, Western University, London, ON, N6A 5B7, Canada. e-mail: ahaque32@uwo.ca}
        
		\thanks{G. Sidebottom is with the Juniper Networks, Kanata, ON, Canada. e-mail:  gsidebot@juniper.net}
}

\markboth{Journal of \LaTeX\ Class Files,~Vol.~14, No.~8, August~2021}%
{Shell \MakeLowercase{\textit{et al.}}: A Sample Article Using IEEEtran.cls for IEEE Journals}


\maketitle

\begin{abstract}

Effective internet traffic prediction in smaller ISP networks is challenged by limited data availability. This paper explores this issue using transfer learning and data augmentation techniques with two LSTM-based models, LSTMSeq2Seq and LSTMSeq2SeqAtn, initially trained on a comprehensive dataset provided by Juniper Networks \cite{junipernetworks} and subsequently applied to smaller datasets. The datasets represent real internet traffic telemetry, offering insights into diverse traffic patterns across different network domains. Our study revealed that while both models performed well in single-step predictions, multi-step forecasts were challenging, particularly in terms of long-term accuracy. In smaller datasets, LSTMSeq2Seq generally outperformed LSTMSeq2SeqAtn, indicating that higher model complexity does not necessarily translate to better performance. The models' effectiveness varied across different network domains, reflecting the influence of distinct traffic characteristics. To address data scarcity, Discrete Wavelet Transform was used for data augmentation, leading to significant improvements in model performance, especially in shorter-term forecasts. Our analysis showed that data augmentation is crucial in scenarios with limited data. Additionally, the study included an analysis of the models' variability and consistency, with attention mechanisms in LSTMSeq2SeqAtn providing better short-term forecasting consistency but greater variability in longer forecasts. The results highlight the benefits and limitations of different modeling approaches in traffic prediction. Overall, this research underscores the importance of transfer learning and data augmentation in enhancing the accuracy of traffic prediction models, particularly in smaller ISP networks with limited data availability.

\end{abstract}


\section{Introduction}

\IEEEPARstart{T}{he} exponential growth of internet traffic over the past decade, driven by advancements in network technologies like the Internet of Things (IoT), Industrial IoT (IIoT), 5G, and cloud computing, has been remarkable. According to the Cisco Annual Internet Report \cite{intro1}, the number of internet users is expected to reach approximately 5.3 billion by 2023. This surge necessitates the design of intelligent, self-organizing networks (SONs) capable of adapting to dynamic behaviors and pre-emptive action-taking.

A critical aspect of assessing network load and performance is network traffic analysis \cite{d2019survey}. Precise traffic prediction is vital for effective network management, impacting resource allocation, traffic scheduling, capacity planning, and anomaly detection \cite{intro2}. However, the complexity of real-world internet traffic, characterized by data heterogeneity, anomalous patterns, and data scarcity, poses significant challenges in developing accurate prediction models \cite{intro3}. Traditional statistical techniques like ARIMA \cite{intro4}, SARIMA \cite{medhn2017mobile}, and Holt-Winter \cite{intro5} struggle with the non-linear aspects of internet traffic. But the advent of machine learning and deep learning has helped in improvements in traffic prediction. These methods outperform classical approaches but are heavily reliant on large historical datasets to understand general traffic patterns. 

In the area of network management, the challenge of data scarcity emerges as a key barrier to the development of efficient traffic prediction models \cite{9662277}. While the rapid evolution of network technologies has exponentially increased internet traffic, the ability to accurately predict this traffic is impeded by the lack of extensive and diverse datasets. This limitation restricts the applicability of advanced machine learning and deep learning techniques to solve many problems \cite{alzubaidi2023survey}. The dependancy on substantial historical data for accurate traffic prediction is particularly problematic for larger Internet Service Providers (ISPs) who manage diverse and expansive networks. Our research is thus motivated by the crucial need to overcome these limitations by innovating methods that can effectively predict internet traffic patterns even in the context of limited data. By addressing the challenge of data scarcity, this study aims to initiate approaches that enhance the accuracy and reliability of traffic predictions, thereby significantly contributing to the field of network management and optimization. 

We accomplish this data scarcity challenge via integrating transfer learning in traffic prediction task. It involves applying knowledge from a model trained on a similar task or domain to a new context \cite{pan2010survey}. Our approach involves developing a predictive model using a large dataset for both single-step and multi-step predictions, forming our source domain model. We then apply this model to build prediction models for our target domains, which have limited data to training an individual prediction model. The application of transfer learning is expected to enhance the performance of prediction models in target domains, crucial for improving network resource management and optimization strategies with limited data availability. This study aims to investigate the efficacy of transfer learning in internet traffic time-series prediction. 

There is another challenge in traffic prediction task and that is individual traffic prediction models for different segments of large networks which is essential due to the diversity in traffic patterns and specific requirements of each segment. Each area, such as residential, commercial, or educational, exhibits unique usage patterns and network demands, necessitating separate models for accurate traffic prediction and efficient resource allocation \cite{10354358}. But creating independent models for each segment is resource-intensive and impractical, particularly for larger network systems. Therefore, our proposed transfer learning-based traffic prediction model offers a practical solution by allowing to leverage existing knowledge and models from one domain and apply them to another. This not only simplifies the model development process but also significantly reduces the resources and time required for building and deploying individual models. 

The integration of transfer learning to solve the problem of data scarcity arises an additional challenge in ensuring that the target domain datasets are adequately sized. If the target domain dataset is too small, it might impede effective model training, leading to underfitting and negating the advantages of knowledge transfer through transfer learning. This realization brings us to another crucial aspect of our research: evaluating the necessity and impact of data augmentation techniques in conjunction with transfer learning. When the target domain dataset is not sufficiently large, data augmentation can play a pivotal role in generating synthetic data samples, thereby enhancing the training process. By artificially expanding the size of the target dataset, data augmentation ensures a more comprehensive learning experience for the model, potentially improving its performance and the effectiveness of the knowledge transfer. To address this, we expand our target domain datasets using the Discrete Wavelet Transform (DWT) \cite{shensa1992discrete} data augmentation technique, maintaining the intrinsic patterns in the original traffic data. This expansion aims to bridge the gap between source and target datasets and improve model generalization. 

Our study, therefore, not only explores the efficacy of transfer learning in internet traffic time-series prediction, characterized by unique data features such as seasonality and trends, but also investigates the interaction between transfer learning and data augmentation. Specifically, we aim to determine whether data augmentation is a necessary step in scenarios where the target domain's dataset is limited and how it contributes to improving model performance. The main contributions of this work are:
\begin{itemize}

\item The key contribution of this study is the innovative use of transfer learning to tackle data scarcity in network traffic prediction. Our approach, which applies knowledge from models trained on extensive datasets to target domains with limited data, represents a significant shift from conventional traffic prediction methods. This strategy is especially crucial for ISPs managing diverse networks, where acquiring vast datasets is often impractical.

\item We further extend the usage of transfer learning by developing individualized prediction models for three different target domain. This approach efficiently circumvents the traditional, resource-intensive process of building separate models for each network segment. Our method demonstrates a practical and scalable solution for creating network management strategies, particularly beneficial for expansive network systems.

\item Another key aspect of our research is the integration of data augmentation techniques in conjunction with transfer learning. We explore the use of the Discrete Wavelet Transform (DWT) to augment the size of target datasets, addressing the challenge of underfitting in smaller datasets. This novel approach not only enriches the training process but also significantly boosts model performance, demonstrating the potential of data augmentation in enhancing transfer learning efficacy.
\end{itemize}

The organization of this paper is as follows: Section \ref{Literature Review} delves into the literature review of current traffic prediction using deep learning models. Section \ref{problem_statement} define the problem and section \ref{Methodology} outlines our methodology, including deep transfer learning, model explanations, and experimental details. Section \ref{result} compares the performance of various deep learning methods on smaller datasets, employing both standard learning and transfer learning. Finally, Section \ref{conclusion} concludes the paper and discusses future research directions.

\section{Literature Review}
\label{Literature Review}
In this section, we review the related work on internet traffic prediction and transfer learning techniques, which are the two main components of our work. 
\subsection{Internet Traffic Prediction}
Recent research has delved into various methods for predicting internet traffic, broadly categorizing them into linear, nonlinear, and hybrid model types.

Linear traffic forecasting models rely on polynomial fitting to understand trends in historical network data for future predictions. While these models are efficient for short-term forecasting, their linear nature makes them unsuitable for medium or long-term predictions, as real-world internet traffic often exhibits nonlinear, periodic, and random characteristics. For instance, the self-similar and long-range dependent nature of internet traffic challenges the accuracy of ARMA (AutoRegressive Moving Average) methods \cite{yang2021network}. Consequently, nonlinear models like FARIMA (Fractional Autoregressive Integrated Moving Average) and SARIMA (Seasonal Autoregressive Integrated Moving Average) have been developed to better capture both short-range and long-range dependencies in internet traffic \cite{yang2021farima}\cite{sarima}.
Despite their individual strengths, both linear and nonlinear models have limitations, leading to the development of hybrid models that combine multiple forecasting methods. Examples include combining linear and nonlinear models, weight mixture models, and decomposition combined models. One such hybrid approach combines FARIMA or FARIMA/GARCH with neural networks for handling both long-term and short-term traffic correlations, although this method can be computationally demanding \cite{liter2}. Integrating FARIMA with alpha stable distribution has shown high accuracy in Wi-Fi traffic prediction and Quality of Service (QoS) provision, but it struggles with non-stationary traffic, a common occurrence in real-world networks \cite{sheng2020alpha}\cite{christian2021network}.

Neural network-based nonlinear models have gained popularity for their ability to handle complex traffic data. Techniques like Echo State Networks (ESN), Fuzzy Neural Networks (FNN), and Radial Basis Function Neural Networks (RBF) have been applied for traffic prediction. ESNs, known for their robust nonlinear capabilities and efficient short-term memory, face limitations in their fixed reservoir design, which can result in sub-optimal performance with diverse traffic data. Adaptive reservoir ESNs (ESN-AR) have been proposed to address this, but they introduce computational complexity \cite{8936255}. FNN models using back-propagation address dynamic mapping in internet traffic forecasting but suffer from slow learning and the risk of getting trapped in local minima \cite{li2019network}. Although FNN-based networks are more accurate, their high computational demand hinders scalability. Traditional RBF networks, using Gradient descent for parameter optimization, struggle with slow convergence and the risk of local optima, leading to extensive computational needs and sub-optimal parameterization, thereby limiting their application in network traffic prediction.

In conclusion, traditional linear models, while effective for short-term forecasting, fail to accommodate the nonlinear, periodic, and random nature of internet traffic, leading to inaccuracies in medium and long-term predictions. Nonlinear models like FARIMA and SARIMA, despite their enhanced capabilities, still struggle with the complexities of real-world internet traffic, particularly in non-stationary scenarios. Hybrid models offer a promising direction but often at the cost of increased computational complexity. Neural network-based nonlinear models, including ESNs, FNNs, and RBFs, have shown potential in handling complex traffic data but are not without their limitations, such as fixed design parameters, computational inefficiency, and scalability issues. These challenges highlight the need for innovative approaches that can effectively predict internet traffic patterns in the context of rapidly evolving network technologies and increasing data heterogeneity. One of the common limitations of existing internet traffic prediction models, particularly in scenarios where it is challenging to collect sufficient data for model training. Most existing models do not adequately address situations with insufficient data for training, nor do they consider the impracticality of developing individual models for different segments of large networks. This oversight in current methodologies often leads to sub-optimal traffic forecasting, especially for expansive and diverse network systems managed by larger ISPs.

\subsection{Internet Traffic Prediction Based on Transfer Learning}
Our research fills a critical gap in traffic prediction by incorporating transfer learning, offering an innovative approach to address data scarcity and the need for segment-specific models in extensive networks. Our method, which capitalizes on established knowledge in one domain to enhance predictions in another with limited data, marks a substantial improvement over traditional techniques. Although some existing studies have integrated transfer learning with traffic prediction models, gaps remain.

For instance, Wu, Qiong, et al. \cite{1} introduced a mobile traffic prediction framework merging parameter-transfer \cite{7} and domain adaptation \cite{8} techniques from deep transfer learning to boost model performance with smaller datasets. This framework involves constructing a target prediction model using a large dataset and then applying pre-trained model knowledge from a data-scarce source domain. They also employed a GAN-based method to address domain shift issues caused by differing data distributions between the source and target domains. While their approach showed improved prediction for smaller datasets, the quality and stability of the GAN-generated data remain unclear. Similarly, Li, Ning, et al. \cite{2} developed a satellite traffic prediction model using a GRU architecture, enhanced by transfer learning and a particle filter algorithm. This approach reduced training time and dataset size requirements. However, their model's performance in scenarios with asymmetric source and target data distributions is not addressed, which is a common real-world challenge. Zeng, Qingtian, et al. \cite{3} proposed a wireless cellular traffic prediction model using cross-domain datasets, optimizing model accuracy by adjusting transferred parameters or features. Their results demonstrated the superiority of models incorporating transfer learning over those without it. Dridi, Aicha, et al. \cite{4}, focusing on time series classification and prediction, implemented transfer learning for improved prediction with smaller datasets and model re-adaptation for different domains. Their results favored transfer learning in time-series prediction, but the applicability to varied real-world data distributions was not explored. Huang, Yunjie, et al. \cite{5} introduced a spatial-temporal traffic prediction model with transfer learning for urban traffic in small to medium-sized cities. They utilized graph neural networks and GRUs to model spatial and temporal traffic data, respectively. Their experiments, using distinct source and target datasets, highlighted transfer learning's benefits in smaller dataset predictions. 


In conclusion, it becomes evident that while there have been significant advances in integrating transfer learning with traffic prediction models, key gaps persist in the existing research, particularly in handling multiple target domains with distinct data distributions and in managing scenarios with inadequate data sizes for deep learning models. Most existing studies have not fully explored the implications of transfer learning across multiple target domains where the data distribution significantly differs from the source domain. This oversight raises questions about the adaptability and robustness of these models in varied network environments, a critical consideration for ISPs managing complex and dynamic networks. Moreover, there is a notable absence of research investigating the effectiveness of transfer learning when the target domain lacks sufficient data to train deep learning models effectively. This scenario is particularly relevant given the increasing instances of data scarcity in various network segments.

Our research aims to address these challenges by conducting a thorough investigation into the application of transfer learning in traffic prediction, especially in situations where target domains have limited data availability and exhibit diverse data distributions compared to the source domain. In essence, our study seeks to bridge the identified gaps in current traffic prediction methodologies by offering a comprehensive and practical solution through transfer learning. This research not only has the potential to advance the field of network traffic prediction but also sets a precedent for future studies to explore the multifaceted applications of transfer learning in diverse and challenging data environments.

\section{Problem and Definitions}
\label{problem_statement}

This study addresses the classic multi-step time-series forecasting challenge: predicting future internet traffic using historical data. We focus on the feasibility of applying transfer learning across two distinct domains. The source domain $S_d$, possesses a dataset denoted as \(X_A = \{x_{A1}, x_{A2}, \ldots, x_{AT}\}\), where \(x_{At}\) represents the traffic at time step \(t\), with \(T\) indicating the total number of time steps. The task for Domain A involves forecasting internet traffic for future steps \(n\), where \(n \in \{6, 9, 12\}\). Therefore, we define the problem for Domain A as: Given the input sequence \(x_{A(t-n_{\text{past}}+1)}, \ldots, x_{At}\), predict the future output sequence \(x_{A(t+1)}, \ldots, x_{A(t+n)}\).

Upon developing a validated model using Domain A's data, we then apply transfer learning to adapt this model for smaller target domains $T_{d1}$, $T_{d2}$, and $T_{d3}$. The aim is to fine-tune the pre-trained model from Domain $S_d$ to predict internet traffic in these target domains. We reformulate the forecasting problem for each target domain as follows: Given the input sequence from the target domain \(x_{T_{d1}(t-n_{\text{past}}+1)}, \ldots, x_{T_{d1}t}\) (and similarly for $T_{d2}$ and $T_{d3}$), we adjust the model to predict the output sequence \(x_{T_{d1}(t+1)}, \ldots, x_{T_{d1}(t+n)}\) (and likewise for $T_{d2}$ and $T_{d3}$).

To tackle data limitations in the target domains and to optimize transfer learning performance, this research introduces data augmentation as a strategic complement. Specifically, we employ Discrete Wavelet Transform (DWT) techniques to artificially expand the datasets of these domains. This expansion is intended to enrich the training base of the models, enhancing their generalization capabilities and the accuracy of multi-step forecasts. We hypothesize that the integration of transfer learning and data augmentation can overcome the data scarcity challenge, thereby fostering the development of robust prediction models for smaller ISP networks.

The effectiveness of this combined approach will be evaluated based on improvements in forecast accuracy and model robustness across the different domains. Ultimately, this study aims to establish a systematic framework for determining the minimum dataset size in target domains that can still benefit from transfer learning, supplemented by data augmentation. This offers a novel approach to resolving a widespread issue in time series forecasting for smaller ISP networks.

\section{Methodology}
\label{Methodology}
In this section, we begin with an overview of our proposed framework in subsection \ref{overview}, highlighting our innovative use of LSTM networks, including an attention-based model. Next, we delve into the specifics of the LSTM Encoder-Decoder model and its attention variant, explaining their roles in capturing and forecasting time series data respectively in subsection \ref{lstmseq2seq} and \ref{lstmseq2sqatn}. We then discuss the application of Discrete Wavelet Transform for data augmentation, aiming to enhance model performance in scenarios with limited data in subsection \ref{dwt}. This is followed by an exploration of transfer learning techniques, adapting a pre-trained model to our specific traffic prediction context in subsection \ref{dtl}. Evaluation metrics such as WAPE, MAE, and RMSE are detailed in subsection \ref{metric} for assessing model performance. Finally, we conclude with the software and hardware preliminaries in subsection \ref{software} used in our experiments.

\begin{figure}
    \centering
    \includegraphics[width=9cm, height=11cm]{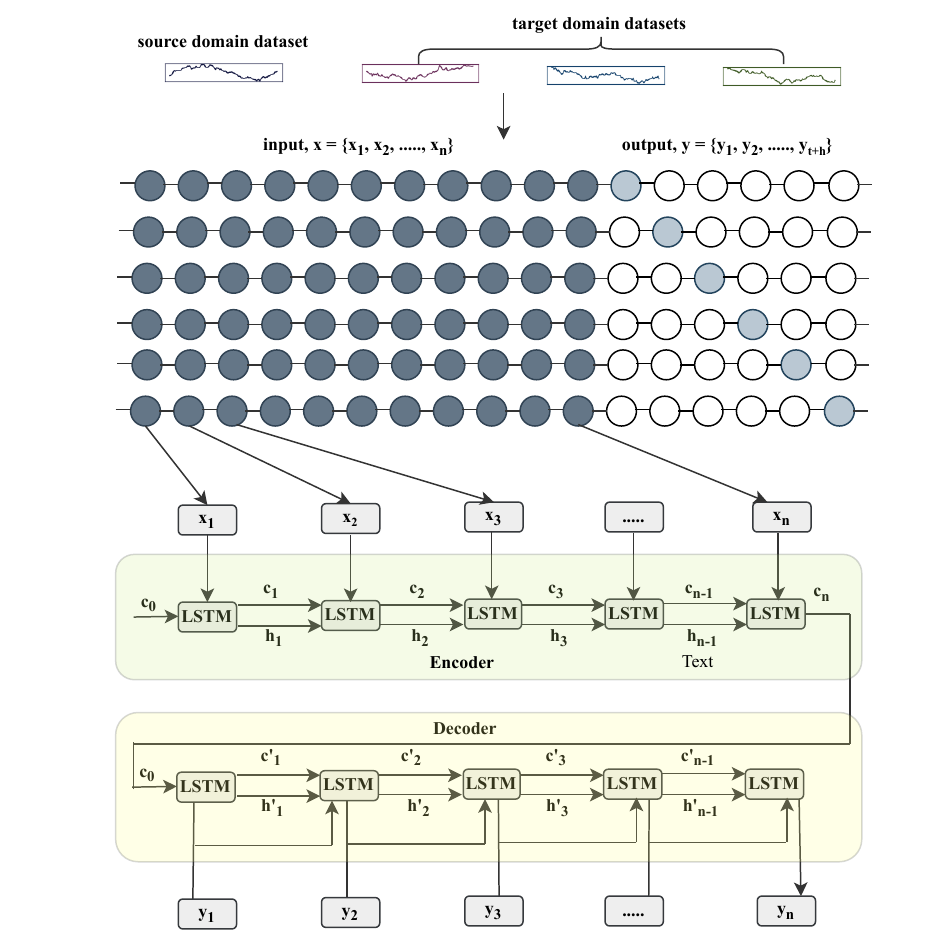}
    \caption{Architecture of the LSTM-based Sequence-to-Sequence Model for Source and Target Domain Internet Traffic Prediction.}
    \label{fig:model}
\end{figure}

\subsection{Overview of the Proposed Framework}
\label{overview}
This study presents two multi-step forecasting approach for univariate time series analysis, incorporating an encoder–decoder structure with one model including an attention layer. This is illustrated in Figure 1. The attention-based model is composed of three principal components: a Long Short-Term Memory (LSTM) network serving both as the encoder and the decoder, and an integrated temporal attention context layer. In contrast, the secondary model operates without an attention layer.

The LSTM encoder in the attention-based framework is designed to capture and interpret hidden patterns within input data of varying sequences, effectively extracting long temporal dependency characteristics inherent in univariate time series data. This is achieved through the temporal attention layer, which helps in the construction of latent space variables, termed temporal attention context vectors. These vectors are pivotal in encapsulating the dynamic temporal relationships within the data. On the other side, the LSTM decoder forecasts future values in the time series. This prediction process leverages the derived latent space variables, ensuring a more informed and accurate forecast by considering the intricate temporal dynamics captured by the attention mechanism.

In this paper, we investigate the efficacy of transfer learning in addressing the challenge of data scarcity in predictive modeling. Our approach involves leveraging knowledge from a well-established source domain to construct predictive models for target domains where data availability is limited. To further enhance this process, we have incorporated a Discrete Wavelet Transform (DWT) based data augmentation technique. This method aims to artificially expand the dataset in the target domain, thereby mitigating the issue of insufficient data samples.

We conduct a comprehensive analysis to assess the impact of the size of the data samples in the target domain on the performance of the transfer learning models. Our study seeks to determine the optimal balance between the quantity of augmented data and the effectiveness of the transfer learning process. This involves examining how well the model can apply what it learned from the source domain (where we have ample data) to the target domain (where data is scarce).

Our findings contribute valuable insights into the optimal strategies for implementing transfer learning in scenarios of data scarcity, particularly in terms of balancing data augmentation with model performance. The results of this study have implications for a wide range of applications where data is limited, offering a pathway to more effective and efficient predictive modeling in such contexts.

\subsection{LSTM Encoder-Decoder (LSTMSeq2Seq) model}
\label{lstmseq2seq}
The Encoder-Decoder LSTM algorithm for time series prediction is a deep learning technique that leverages the power of recurrent neural networks to effectively capture temporal dependencies in time series data. The model consists of two key components: an encoder LSTM and a decoder LSTM. The encoder LSTM processes the input sequence of past time steps and generates a final hidden state and cell state, which encapsulate the relevant information from the input. This final hidden state is then replicated and used as the initial input for the decoder LSTM. 

\begin{figure}
    \centering
    \includegraphics[width=7cm, height=6cm]{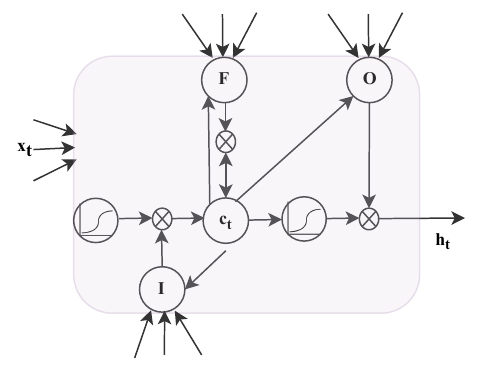}
      \caption{Detailed Structure of a Long Short-Term Memory (LSTM) Cell at Time Step $t$.}
    \label{fig:lstm_cell}
\end{figure}

An LSTM (Long Short-Term Memory) unit depicted in Figure \ref{fig:lstm_cell} is a type of recurrent neural network (RNN) cell that is adept at capturing long-term dependencies in sequential data. It does so through a series of gates. The forget gate decides what information should be thrown away from the cell state. It looks at the previous hidden state $h_{t-1}$ and the current input $x_t$, and outputs a number between 0 and 1 for each number in the cell state $c_{t-1}$. An LSTM cell consist of cell state $(C)$, hidden state $(h)$, and three gates called input gate$(i)$, forget gate$(f)$, and output gate $(o)$

The forget gate decides what information should be thrown away from the cell state. It looks at the previous hidden state $h_{t-1}$ and the current input $x_t$, and outputs a number between 0 and 1 for each number in the cell state 
$C_{t-1}$.

\begin{equation}
  f_t = \sigma(W_f \cdot [h_{t-1}, x_t] + b_f) 
\end{equation}
    
The input gate decides which values will be updated and a candidate layer creates a vector of new candidate values that could be added to the cell state.

\begin{equation}
    i_t = \sigma(W_i \cdot [h_{t-1}, x_t] + b_i)
\end{equation}
\begin{equation}
    \tilde{C}_t = \tanh(W_C \cdot [h_{t-1}, x_t] + b_C) 
\end{equation}

Finally, the output gate decides what the next hidden state should be. The cell state is passed through tanh (to normalize it between -1 and 1) and multiplied by the output of the output gate, so that the hidden state only contains the parts we decided to output.

\begin{equation}
    o_t = \sigma(W_o \cdot [h_{t-1}, x_t] + b_o)
\end{equation}
\begin{equation}
   h_t = o_t * \tanh(C_t) 
\end{equation}

The LSTM encoder processes the input sequence and generates a hidden state $h_t$ and a cell state $c_t$ at the final time step $t$. The LSTM decoder then processes these states to generate the output sequence. Let $f_{enc}$ and $f_{dec}$ represent the encoder and decoder functions, respectively. The prediction problem can be further represented as:

\begin{equation}
    (h_t, c_t) = f_{enc}(x_{t-n_{past}+1}, x_{t-n_{past}+2}, ..., x_t)
\end{equation}
\begin{equation}
    (x_{t+1}, x_{t+2}, ..., x_{t+n}) = f_{dec}(h_t, c_t)
\end{equation}

To obtain the final prediction, a TimeDistributed dense layer is applied to the output sequence of the decoder LSTM. The model is compiled using the Adam optimizer and the Huber loss function and trained on the given training data. This approach effectively captures the temporal dynamics of the input sequence and allows for accurate predictions of future time steps in the time series data.

\begin{figure}
    \centering
    \includegraphics[width=5.5cm, height=4.5cm]{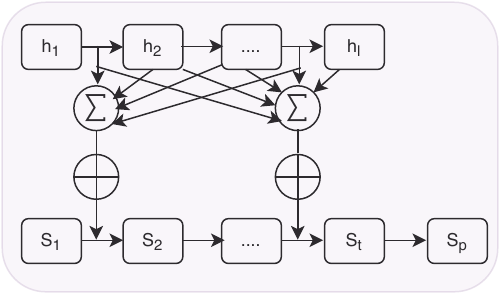}
        \caption{High-level Diagram of Attention Mechanism in a Seq2Seq Model.}
    \label{fig:attention}
\end{figure}

The Encoder-Decoder LSTM algorithm for time series prediction uses various parameters that contribute to the model's performance. The choice of 100 hidden units for both the encoder and decoder LSTM layers balances model complexity and computational efficiency. A higher number of hidden units would increase the capacity to capture complex patterns in the data but could lead to overfitting and longer training times. Conversely, fewer hidden units might reduce the model's ability to learn and represent the underlying structure of the time series data. The use of the Adam optimizer is motivated by its adaptive learning rate, which adjusts based on the gradient's magnitude, resulting in faster convergence compared to traditional gradient descent methods. This makes it suitable for training deep neural networks such as LSTM models. The Huber loss function is employed as it combines the best properties of Mean Squared Error (MSE) and Mean Absolute Error (MAE) loss functions. It is less sensitive to outliers than MSE and has a smoother gradient than MAE, allowing for a more stable learning process. The TimeDistributed dense layer is applied to the output sequence of the decoder LSTM to obtain the final prediction, enabling the model to process each time step independently while maintaining the same weights across all time steps. This approach helps the model generalize to varying sequence lengths and makes it more robust to the intricacies of time series data. These parameter choices, while not exhaustive, provide a solid foundation for the Encoder-Decoder LSTM model. Further fine-tuning or hyperparameter optimization could be performed based on the specific dataset and problem domain to improve the model's performance.

\begin{algorithm}
\caption{Encoder-Decoder LSTM for Time Series Prediction}
\begin{algorithmic}[1]
\STATE Initialize: \( \mathcal{N}_{future} = \{n_{future}\}, \mathcal{N}_{past} = \{n_{past}\} \)
\STATE \( \text{InputSeq}(\mathcal{N}_{past}, \mathcal{N}_{future}) \)
\STATE \( \text{SplitData}(\text{InputData}) \rightarrow \text{Data}_{train}, \text{Data}_{test} \)
\FOR{\( n_{future} \in \mathcal{N}_{future} \)}
    \FOR{\( n_{past} \in \mathcal{N}_{past} \)}
        \STATE \( \text{InputSeq}(n_{past}, n_{future}) \)
        \STATE \( \text{SplitData}(\text{InputData}) \rightarrow \text{Data}_{train}, \text{Data}_{test} \)
        \STATE Encoder LSTM: \( \text{LSTM}_{enc}^{100} \)
        \STATE \( h_t = \text{LSTM}_{enc}(x_t, h_{t-1}), \quad t = 1, \ldots, T \)
        \STATE \( h_T, c_T \) (final states)
        \STATE \( \text{decoder\_input} = \text{RepeatVector}(n_{future})(h_T) \)
        \STATE Decoder LSTM: \( \text{LSTM}_{dec}^{100}(h_T, c_T) \)
        \STATE \( y_t = \text{LSTM}_{dec}(y_{t-1}, h_{t-1}), \quad t = 1, \ldots, n_{future} \)
        \STATE \( y = (y_1, \ldots, y_{n_{future}}) \)
        \STATE \( y' = \text{TimeDistributed}(\text{Dense}(1))(y) \)
        \STATE Compile: \( \text{Adam optimizer}, \text{Huber loss} \)
        \STATE Train and evaluate on \( \text{Data}_{train}, \text{Data}_{test} \)
    \ENDFOR
\ENDFOR
\end{algorithmic}
\end{algorithm}

\subsection{LSTM Encoder-Decoder Model with Attention Layer (LSTMSeq2SeqAtn)} 
\label{lstmseq2sqatn}

The LSTM encoder-decoder model with attention is designed to enhance the decoder's ability to selectively focus on different parts of the input sequence for each step of the output generation. This attention mechanism depicted in Figure \ref{fig:attention} dynamically calculates a set of attention weights, $a_t$, that represent the relevance of each input time step to the current output. 

The encoder function, $f_{enc}$, processes the input sequence to produce a sequence of hidden states $(h_1, h_2, \ldots, h_t)$ and a cell state $c_t$ as shown in Equation (8). The attention function, $f_{att}$, then computes the attention weights by taking the dot product between the decoder's current state and each of the encoder's hidden states, as depicted in Equation (9) and visually represented in the attached image, where the sum symbols and directed lines correspond to the computation of attention weights.

\begin{equation}
    (h_t, c_t) = f_{enc}(x_{t-n_{past}+1}, x_{t-n_{past}+2}, ..., x_t)
\end{equation}

\begin{equation}
    a_t = f_{att}(h_t, h_{t-1}, c_{t-1})
\end{equation}

\begin{equation}
    (x_{t+1}, x_{t+2}, ..., x_{t+n}) = f_{dec}(h_t, c_t, a_t)
\end{equation}

These weights are then applied to the encoder's hidden states to create a context vector, which encapsulates the focused information from the input sequence relevant to the current output time step. The context vector is computed as a weighted sum of the encoder hidden states, where the weights are derived from the attention scores, emphasizing parts of the input sequence that are more pertinent to the current decoder state. The decoder function, $f_{dec}$, utilizes this context vector along with the decoder's current state to generate the subsequent output sequence $(x_{t+1}, x_{t+2}, ..., x_{t+n})$, as per Equation (10). This process is iteratively applied to each time step in the output sequence, with the attention mechanism recalculating the context vector at each step, thus allowing the decoder to produce a sequence that is conditioned on selective parts of the input.

During the initialization phase, the input sequences are prepared with respect to $n_{past}$ and $n_{future}$, the past and future time steps respectively. This preparation is crucial as it determines the temporal context for the model. The encoder LSTM layer, with a chosen 100 hidden units, captures the temporal dependencies within the input sequence. The final hidden state, $h_t$, is then replicated and used to initialize the decoder LSTM layer, also with 100 hidden units, which in turn processes the decoder input and generates predictions. The attention scores between encoder and decoder hidden states are calculated at each time step, forming the basis for the context vector. The combination of the context vector with the decoder hidden states is then used by a TimeDistributed dense layer to make the final prediction, $y'$. The model is compiled with the Adam optimizer, known for its efficiency in handling sparse gradients and adaptive learning rate, and the Huber loss function, which is less sensitive to outliers in data than other loss functions. The model is then trained on the training data set, adjusting weights through backpropagation, and its performance is evaluated on the testing data set for each combination of $n_{past}$ and $n_{future}$.

\begin{algorithm}
\caption{Encoder-Decoder LSTM with Attention for Time Series Prediction}
\begin{algorithmic}[1]
\STATE Initialize: \( \mathcal{N}_{future} = \{n_{future}\}, \mathcal{N}_{past} = \{n_{past}\} \)
\STATE \( \text{InputSeq}(\mathcal{N}_{past}, \mathcal{N}_{future}) \)
\STATE \( \text{SplitData}(\text{InputData}) \rightarrow \text{Data}_{train}, \text{Data}_{test} \)
\FOR{\( n_{future} \in \mathcal{N}_{future} \)}
    \FOR{\( n_{past} \in \mathcal{N}_{past} \)}
        \STATE Prepare tensors: \( \text{Data}_{train}, \text{Data}_{test} \)
        \STATE Encoder LSTM: \( \text{LSTM}_{enc}^{100} \)
        \STATE \( h_t = \text{LSTM}_{enc}(x_t, h_{t-1}), \quad t = 1, \ldots, T \)
        \STATE \( h_T, c_T, h = (h_1, \ldots, h_T) \)
        \STATE Decoder init: \( \text{Input}_{dec} = \text{Replicate}(h_T, n_{future}) \)
        \STATE Decoder LSTM: \( \text{LSTM}_{dec}^{100}(h_T, c_T) \)
        \STATE \( y_t = \text{LSTM}_{dec}(y_{t-1}, h_{t-1}), \quad t = 1, \ldots, n_{future} \)
        \STATE Attention scores: \( \text{Attention} = \text{softmax}(\text{dot}(\text{decoder\_stack\_h}, \text{encoder\_stack\_h}, \text{ax}
        =[2, 2])) \)
        \STATE Context vector: \( \text{context} = \text{dot}(\text{Attention}, \text{encoder\_stack\_h}, \text{ax}=[2,1]) \)
        \STATE \( \text{decoder\_combined\_context} = \text{concatenate}(\text{context}, \text{decoder\_stack\_h}) \)
        \STATE \( y' = \text{TimeDistributed}(\text{Dense}(1))(\text{decoder\_context})\)
        \STATE Compile: \( \text{Adam optimizer}, \text{Huber loss} \)
        \STATE Train and evaluate on \( \text{Data}_{train}, \text{Data}_{test} \)
    \ENDFOR
\ENDFOR
\end{algorithmic}
\end{algorithm}

\subsection{Data Augmentation Using Discrete Wavelet Transformation}
\label{dwt}
Discrete Wavelet Transform (DWT) is a data augmentation technique employed in this study to address the challenge of limited data in the target domains, which can significantly impact the performance of machine learning models \cite{iwana2021time}. DWT is a powerful tool for time-frequency analysis and has been widely used for signal processing, image compression, and feature extraction \cite{mallat2009wavelet}. By augmenting the dataset with new, meaningful information using DWT, this study aims to enhance the performance of the LSTMSeq2Seq and LSTMSeq2SeqAtn models by providing a richer and more diverse representation of the target domains. The process of applying the DWT technique to the smaller datasets of the target domains involves several steps:

\begin{enumerate}
\item \textbf{Wavelet Decomposition:} The original data is decomposed using a wavelet function. Let $x(t)$ be the original signal. The wavelet decomposition at level $j$ can be expressed as:
    \begin{equation}
        x(t) = \sum_{k} c_{j,k} \phi_{j,k}(t) + \sum_{k} d_{j,k} \psi_{j,k}(t)
    \end{equation}
    where $c_{j,k}$ are approximation coefficients, $d_{j,k}$ are detail coefficients, and $\phi_{j,k}(t)$, $\psi_{j,k}(t)$ are the scaling and wavelet functions at level $j$ and position $k$, respectively.

\item \textbf{Modifying the Detail Coefficients:} The detail coefficients $d_{j,k}$ from higher frequency bands are modified. Each detail coefficient is multiplied by a random factor $\alpha$ within a specified range (e.g., 0.5 to 1.5):
    \begin{equation}
        d'_{j,k} = \alpha \times d_{j,k}
    \end{equation}
    This introduces variability while retaining the core structure of the original data.
    
\item \textbf{Inverse Wavelet Transform:} An inverse wavelet transform is applied to the modified coefficients to reconstruct data points. The reconstructed signal $x'(t)$ is given by:
    \begin{equation}
        x'(t) = \sum_{k} c_{j,k} \phi_{j,k}(t) + \sum_{k} d'_{j,k} \psi_{j,k}(t)
    \end{equation}

\item \textbf{Combining Data:} The original and augmented data are combined. The final dataset $D$ is a union of the original dataset $D_{\text{orig}}$ and the augmented dataset $D_{\text{aug}}$:
    \begin{equation}
        D = D_{\text{orig}} \cup D_{\text{aug}}
    \end{equation}

\end{enumerate}

By employing the DWT data augmentation technique, the study anticipates improving the performance of the LSTMSeq2Seq and LSTMSeq2SeqAtn models by enabling them to learn more diverse and robust representations of the target domains. This enhancement is expected to lead to better generalization and reduced overfitting when tested on smaller datasets.

\begin{algorithm}
\caption{Discrete Wavelet Transform (DWT) Data Augmentation}
\begin{algorithmic}[1]
\STATE \textbf{Input:} Original dataset $D_{orig}$, wavelet function $w$ (e.g., 'db4'), number of levels $l$, augmentation factor range $[a, b]$ (e.g., $[0.5, 1.5]$)
\STATE \textbf{Output:} Augmented dataset $D_{aug}$
\FOR{each data point $x_i$ in $D_{orig}$}
    \STATE Perform wavelet decomposition of $x_i$ using $w$ and $l$ to obtain sets of coefficients $C_i = \{c_{i1}, c_{i2}, ..., c_{in}\}$
    \FOR{each set $c_{ij}$ in $C_i$ corresponding to higher frequency bands}
        \STATE $c_{ij}' = c_{ij} \times f_{ij}$ where $f_{ij} \sim Uniform(a, b)$
    \ENDFOR
    \STATE Perform inverse wavelet transform on modified coefficients $C_i'$ to generate new data point $x_i'$
    \STATE $D_{aug} = D_{aug} \cup \{x_i'\}$
\ENDFOR
\STATE $D_{exp} = D_{orig} \cup D_{aug}$
\RETURN $D_{exp}$
\end{algorithmic}
\end{algorithm}

\begin{algorithm}
\caption{Transfer Learning for Time Series Prediction}
\begin{algorithmic}[1]
\STATE Initialize: \( \mathcal{N}_{future} = \{n_{future}\}, \mathcal{N}_{past} = \{n_{past}\} \)
\STATE \( \text{InputSeq}(\mathcal{N}_{past}, \mathcal{N}_{future}) \)
\STATE \( \text{SplitData}(\text{InputData}) \rightarrow \text{Data}_{train}, \text{Data}_{test} \)
\FOR{\( n_{future} \in \mathcal{N}_{future} \)}
    \FOR{\( n_{past} \in \mathcal{N}_{past} \)}
        \STATE \( \text{InputSeq}(n_{past}, n_{future}) \)
        \STATE \( \text{SplitData}(\text{InputData}) \rightarrow \text{Data}_{train}, \text{Data}_{test} \)
        \STATE \( M_s \leftarrow \text{LoadModel}() \)
        \STATE \( M_s \leftarrow M_s \setminus \{D_s\} \)
        \STATE \( M_s \leftarrow M_s \cup \{D_t\} \)
        \STATE \( M_t = M_s \cup \{D_t\} \)
        \STATE \( \text{FreezeLayers}(M_t \setminus \{D_t\}) \)
        \STATE \( \text{Compile}(M_t, \text{Adam}(0.001), \text{Huber}) \)
        \STATE \( \text{Train}(M_t, D^{(T)}_{train}, \text{epochs}) \)
        \STATE \( \text{UnfreezeLayers}(M_t) \)
        \STATE \( \text{Compile}(M_t, \text{Adam}(0.0001), \text{Huber}) \)
        \STATE \( \text{FineTune}(M_t, D^{(T)}_{train}, \text{epochs}) \)
        \STATE \( \text{Evaluate}(M_t, D^{(T)}_{test}) \)
    \ENDFOR
\ENDFOR
\end{algorithmic}
\end{algorithm}

\subsection{Proposed Transfer Learning Based Traffic Prediction Model}
\label{dtl}
Transfer learning is an deep learning or machine learning optimization approach in which knowledge is transferred from one domain to another similar domain. We can formally define transfer learning in terms of domain and task \cite{11}. There are two domains such as source and target domain involved in transfer learning while the domain consists of a feature space $X$ and probability distribution $P(X)$ where $X=\{x_1, x_2,..., x_n\}\in X$. The task $T$ in transfer learning is consists of label space $Y$ and an objective function $f:X \longrightarrow Y$ while $X$ is a feature space for particular domain, $\{X, P(X)\}$. In transfer learning, the source domain $D_S$ and target domain $D_T$ consists of two different task $T_S$ and $T_T$ and the purpose of the transfer learning is to assist the task in target domain to perform better using the knowledge in $D_S$ and $T_S$.  

In this study, we employ transfer learning to adapt a pre-trained Encoder-Decoder LSTM model from a source domain to a target domain, particularly when the target domain dataset is relatively small. The rationale behind this approach is to leverage the knowledge acquired by the model on the source domain, where abundant data is available, to improve its performance on the target domain with limited data. This technique reduces the risk of overfitting and allows the model to generalize better to the target domain. To implement the transfer learning process, we first remove the output layer of the pre-trained source model, which is tailored to the source domain's prediction task, and replace it with a new dense layer designed for the target domain. This modification ensures that the output of the adapted model is compatible with the target domain's prediction task. During the initial training phase, we freeze all layers of the model except the newly added dense layer by setting their trainable attribute to False. This approach allows the model to focus on learning the target domain-specific characteristics without altering the weights of the other layers, which have already captured useful information from the source domain. We use a higher learning rate of 0.001 during this phase to encourage faster convergence. Subsequently, we unfreeze all layers in the model and fine-tune it on the target domain data with a reduced learning rate of 0.0001. This lower learning rate ensures that the fine-tuning process makes small, precise updates to the model's weights, allowing it to adapt to the target domain while preserving the knowledge acquired from the source domain. Throughout the training process, we employ the Adam optimizer and the Huber loss function, which have been proven effective in training deep neural networks, such as LSTM models. By utilizing transfer learning in combination with the Encoder-Decoder LSTM model, we effectively adapt the model to the target domain while leveraging the knowledge gained from the source domain. This approach results in improved performance on the smaller target domain dataset, demonstrating the potential benefits of transfer learning for time series prediction tasks in various domains.

\subsection{Evaluation Metrics}
\label{metric}
We used several metrics to estimate the performance of our traffic forecasting models, including Weighted Absolute Percentage Error (WAPE), Mean Absolute Error (MAE), and Root Mean Square Error (RMSE). These performance metrics identify the deviation of the predicted result from the original data and provide a comprehensive understanding of the model's accuracy.

\begin{equation}
    WAPE = \frac{\sum_{i=1}^{n} |p_i - o_i|}{\sum_{i=1}^{n} |o_i|} \times 100\%
\end{equation}

\begin{equation}
    MAE = \frac{1}{n} \sum_{i=1}^{n} |p_i - o_i|
\end{equation}

\begin{equation}
    RMSE = \sqrt{\frac{1}{n} \sum_{i=1}^{n} (p_i - o_i)^2}
\end{equation}

Here, $p_i$ and $o_i$ are the predicted and original values, respectively, and $n$ is the total number of test instances.

\subsection{Software and Hardware Preliminaries}
\label{software}
We used Python and deep learning library TensorFlow-Keras\cite{17} to conduct the experiments.  Our computer has the configuration of Intel (R) i5-9500T CPU@2.20GHz, 8GB memory, and a 64-bit Windows operating system.

\section{Result Analysis and Discussion}
\label{result}

\subsection{Dataset Description}
We gratefully acknowledge Juniper Networks \cite{junipernetworks} for sharing the datasets utilized in our experiments. The data represents real internet traffic telemetry collected from several high-speed interfaces. This telemetry was obtained by sampling the SNMP (Simple Network Management Protocol) interface MIB (Management Information Base) counters on a core-facing interface of a provider edge router. Sampling was conducted at 5-minute intervals, with the bps (bits per second) value for each interval calculated by the difference in samples, multiplied by eight. The interface, with a capacity of 40 Gbps, showed no discards during the sampling period, ensuring the integrity of traffic measurement. Our experiment incorporated a total of four datasets: one source domain dataset $D_s$ for building the baseline predictive model, and three smaller target domain datasets $D_{t1}$,  $D_{t2}$, and $D_{t3}$, each pertaining to traffic from specific source-destination node pairs.

Transfer learning was employed in this experiment to address the challenge of limited data availability in the target domain and to leverage the knowledge learned from the larger source domain dataset to improve predictions in the target domain. The source and target domain datasets share similar attributes, such as timestamps and corresponding internet traffic volume at specific time-points. However, there are some key differences between the domains that could impact the transfer learning performance. First, the data distributions differ between the source and target domains. The source domain dataset $D_s$ has a considerably higher mean traffic volume ($8,364,386,961$) compared to the target domain datasets $D_{t1}$ ($552,753,735.6$), $D_{t2}$ ($729,400,375.1$), and $D_{t3}$ ($553,417,309.4$). This indicates that the source domain experiences significantly more traffic on average than the target domains. Furthermore, the standard deviation and variance values also exhibit differences between the source and target domains. The source domain dataset $D_s$ has a higher standard deviation ($4,098,702,833$) and variance ($1.68 \times 10^{19}$) compared to the target domain datasets $D_{t1}$ (standard deviation: 210,546,994.7, variance: $4.43 \times 10^{16}$), $D_{t2}$ (standard deviation: $287,908,086.1$, variance: $8.29 \times 10^{16}$), and $D_{t3}$ (standard deviation: $210,166,274.1$, variance: $4.42 \time 10^{16}$). This implies that the source domain exhibits greater variability in traffic volume compared to the target domains.

The skewness values for datasets $D_s$, $D_{t1}$, $D_{t2}$, and $D_{t3}$ reveal differences in the asymmetry of their data distributions. Dataset $D_s$ has a positive skewness, indicating a longer right tail with more instances of higher traffic volume values. Datasets $D_{t1}$ and $D_{t3}$ exhibit slightly negative skewness, suggesting somewhat longer left tails with more instances of lower traffic volume values, though the asymmetry is not very strong in either case. Dataset $D_{t2}$ has a skewness value close to 0, indicating a nearly symmetric distribution. These differences in skewness among the datasets could impact the transfer learning performance, as models may need to adapt to varying data asymmetries between the source and target domains. Despite these differences, our transfer learning approach aims to leverage the shared attributes and knowledge gained from the larger source domain dataset to improve predictions in the target domain.


\begin{table*}
\centering
\caption{Summary Statistics Comparison across Different Datasets.}
\label{table_dataset_summary}
\begin{tabular}{l|l|l|l|l} 
\hline
                        & Mean        & STD         & VAR      & Skewness      \\ 
\hline
Source Domain Dataset $D_s$ & 8364386961  & 4098702833  & 1.68E+19 & 0.793776342   \\
Target Domain Dataset $D_{t1}$ & 552753735.6 & 210546994.7 & 4.43E+16 & -0.249849673  \\
Target Domain Dataset $D_{t2}$ & 729400375.1 & 287908086.1 & 8.29E+16 & 0.057702382   \\
Target Domain Dataset $D_{t3}$ & 553417309.4 & 210166274.1 & 4.42E+16 & -0.258100096  \\
\hline
\end{tabular}
\end{table*}

\begin{table}
\centering
\caption{Prediction Model Performance Summary in Source Domain.}
\label{tab:dtl_source_domain_performance}
\begin{tabular}{|c|c|c|} 
\hline
        & \textbf{LSTMSeq2Seq} & \textbf{LSTMSeq2SeqAtn}  \\ 
\hline
        & \textbf{Average Accuracy}       & \textbf{Average Accuracy}            \\ 
\hline
Step 6  & 93.72\%                & 93.62\%                 \\ 
\hline
Step 9  & 92.46\%               & 92.47\%                   \\ 
\hline
Step 12 & 91.20\%                & 91.26\%                     \\
\hline
\end{tabular}
\end{table}

\begin{table*}
\centering
\caption{Comparative Analysis of LSTMSeq2Seq and LSTMSeq2SeqAtn Models in the Target Domain $T_{d1}$ for Forecasting Accuracy Over Different Steps in Two Scenarios: Before and After Data Augmentation. This table evaluates the performance metrics—Mean Absolute Error (MAE), Root Mean Square Error (RMSE), and Weighted Absolute Percentage Error (WAPE)—across forecast horizons of 6, 9, and 12 steps. It provides detailed values for each step and the average performance, highlighting the models' efficacy in predicting Target Domain $T_{d1}$.}
\label{table_target_domain_1}
\begin{tabular}{l|l|lll|lll||lll|lll} 
\hline
                          &                           & \multicolumn{6}{c||}{Target Domain $T_{d1}$ Before Data Augmentation}                                                                                                    & \multicolumn{6}{c}{Target Domain $T_{d1}$ After Data Augmentation}                                                                                                        \\ 
\hline
                          &                           & \multicolumn{3}{c|}{LSTMSeq2Seq}                                                & \multicolumn{3}{c||}{LSTMSeq2SeqAtn}                                            & \multicolumn{3}{c|}{LSTMSeq2Seq}                                                & \multicolumn{3}{c}{LSTMSeq2SeqAtn}                                               \\ 
\hline
Forecast Length           &                           & \multicolumn{1}{c}{MAE} & \multicolumn{1}{c}{RMSE} & \multicolumn{1}{c|}{WAPE}  & \multicolumn{1}{c}{MAE} & \multicolumn{1}{c}{RMSE} & \multicolumn{1}{c||}{WAPE} & \multicolumn{1}{c}{MAE} & \multicolumn{1}{c}{RMSE} & \multicolumn{1}{c|}{WAPE}  & \multicolumn{1}{c}{MAE} & \multicolumn{1}{c}{RMSE} & \multicolumn{1}{c}{WAPE}    \\ 
\hline
\multirow{6}{*}{6 Step}   & Step 1                    & 0.108                   & 0.139                    & 18.452\%                   & 0.102                   & 0.129                    & 17.450\%                   & 0.069                   & 0.099                    & 12.361\%                   & 0.073                   & 0.107                    & 13.095\%                    \\
                          & Step 2                    & 0.098                   & 0.122                    & 16.789\%                   & 0.101                   & 0.130                    & 17.409\%                   & 0.068                   & 0.099                    & 12.191\%                   & 0.076                   & 0.108                    & 13.574\%                    \\
                          & Step 3                    & 0.096                   & 0.124                    & 16.604\%                   & 0.105                   & 0.140                    & 18.234\%                   & 0.069                   & 0.100                    & 12.307\%                   & 0.077                   & 0.109                    & 13.832\%                    \\
                          & Step 4                    & 0.105                   & 0.140                    & 18.138\%                   & 0.107                   & 0.144                    & 18.458\%                   & 0.069                   & 0.099                    & 12.323\%                   & 0.073                   & 0.105                    & 13.159\%                    \\
                          & Step 5                    & 0.103                   & 0.143                    & 17.800\%                   & 0.104                   & 0.141                    & 17.955\%                   & 0.068                   & 0.097                    & 12.234\%                   & 0.075                   & 0.104                    & 13.365\%                    \\
                          & Step 6                    & 0.099                   & 0.136                    & 16.957\%                   & 0.100                   & 0.134                    & 17.123\%                   & 0.068                   & 0.101                    & 12.262\%                   & 0.074                   & 0.105                    & 13.290\%                    \\
                          & \textit{\textbf{Average}} & \textit{\textbf{0.102}} & \textit{\textbf{0.134}}  & \textit{\textbf{17.457\%}} & \textit{\textbf{0.103}} & \textit{\textbf{0.136}}  & \textit{\textbf{17.771\%}} & \textit{\textbf{0.069}} & \textit{\textbf{0.099}}  & \textit{\textbf{12.279\%}} & \textit{\textbf{0.075}} & \textit{\textbf{0.106}}  & \textit{\textbf{13.386\%}}  \\ 
\hline
\multirow{9}{*}{9 Step}   & Step 1                    & 0.106                   & 0.135                    & 17.738\%                   & 0.115                   & 0.154                    & 19.258\%                   & 0.071                   & 0.102                    & 12.611\%                   & 0.073                   & 0.105                    & 13.090\%                    \\
                          & Step 2                    & 0.103                   & 0.131                    & 17.429\%                   & 0.120                   & 0.163                    & 20.412\%                   & 0.068                   & 0.100                    & 12.188\%                   & 0.073                   & 0.104                    & 13.007\%                    \\
                          & Step 3                    & 0.102                   & 0.136                    & 17.524\%                   & 0.116                   & 0.170                    & 19.889\%                   & 0.070                   & 0.100                    & 12.528\%                   & 0.074                   & 0.105                    & 13.262\%                    \\
                          & Step 4                    & 0.104                   & 0.138                    & 17.766\%                   & 0.119                   & 0.170                    & 20.274\%                   & 0.068                   & 0.097                    & 12.216\%                   & 0.072                   & 0.103                    & 12.973\%                    \\
                          & Step 5                    & 0.105                   & 0.138                    & 17.949\%                   & 0.120                   & 0.165                    & 20.556\%                   & 0.069                   & 0.099                    & 12.432\%                   & 0.071                   & 0.099                    & 12.652\%                    \\
                          & Step 6                    & 0.107                   & 0.140                    & 18.322\%                   & 0.122                   & 0.170                    & 20.898\%                   & 0.069                   & 0.097                    & 12.366\%                   & 0.070                   & 0.099                    & 12.591\%                    \\
                          & Step 7                    & 0.107                   & 0.141                    & 18.234\%                   & 0.115                   & 0.163                    & 19.542\%                   & 0.069                   & 0.097                    & 12.310\%                   & 0.072                   & 0.104                    & 12.983\%                    \\
                          & Step 8                    & 0.105                   & 0.139                    & 17.685\%                   & 0.112                   & 0.159                    & 18.862\%                   & 0.070                   & 0.098                    & 12.484\%                   & 0.078                   & 0.110                    & 13.903\%                    \\
                          & Step 9                    & 0.106                   & 0.138                    & 17.872\%                   & 0.111                   & 0.158                    & 18.686\%                   & 0.070                   & 0.098                    & 12.599\%                   & 0.074                   & 0.103                    & 13.243\%                    \\
                          & \textit{\textbf{Average}} & \textit{\textbf{0.105}} & \textit{\textbf{0.138}}  & \textit{\textbf{17.836\%}} & \textit{\textbf{0.117}} & \textit{\textbf{0.164}}  & \textit{\textbf{19.820\%}} & \textit{\textbf{0.069}} & \textit{\textbf{0.099}}  & \textit{\textbf{12.415\%}} & \textit{\textbf{0.073}} & \textit{\textbf{0.104}}  & \textit{\textbf{13.078\%}}  \\ 
\hline
\multirow{13}{*}{12 Step} & Step 1                    & 0.109                   & 0.138                    & 18.052\%                   & 0.104                   & 0.141                    & 17.132\%                   & 0.072                   & 0.102                    & 12.834\%                   & 0.074                   & 0.106                    & 13.256\%                    \\
                          & Step 2                    & 0.108                   & 0.140                    & 18.046\%                   & 0.099                   & 0.147                    & 16.475\%                   & 0.070                   & 0.100                    & 12.512\%                   & 0.074                   & 0.105                    & 13.128\%                    \\
                          & Step 3                    & 0.108                   & 0.146                    & 18.090\%                   & 0.113                   & 0.182                    & 18.992\%                   & 0.070                   & 0.099                    & 12.469\%                   & 0.073                   & 0.104                    & 12.997\%                    \\
                          & Step 4                    & 0.110                   & 0.150                    & 18.346\%                   & 0.116                   & 0.187                    & 19.426\%                   & 0.068                   & 0.096                    & 12.170\%                   & 0.070                   & 0.099                    & 12.508\%                    \\
                          & Step 5                    & 0.111                   & 0.153                    & 18.738\%                   & 0.120                   & 0.189                    & 20.340\%                   & 0.069                   & 0.096                    & 12.300\%                   & 0.071                   & 0.100                    & 12.758\%                    \\
                          & Step 6                    & 0.111                   & 0.154                    & 18.773\%                   & 0.120                   & 0.195                    & 20.216\%                   & 0.069                   & 0.096                    & 12.416\%                   & 0.070                   & 0.099                    & 12.597\%                    \\
                          & Step 7                    & 0.113                   & 0.154                    & 19.003\%                   & 0.118                   & 0.193                    & 19.874\%                   & 0.070                   & 0.099                    & 12.511\%                   & 0.072                   & 0.102                    & 12.816\%                    \\
                          & Step 8                    & 0.112                   & 0.151                    & 18.758\%                   & 0.120                   & 0.188                    & 20.167\%                   & 0.071                   & 0.099                    & 12.696\%                   & 0.072                   & 0.102                    & 12.875\%                    \\
                          & Step 9                    & 0.113                   & 0.154                    & 19.052\%                   & 0.119                   & 0.183                    & 19.981\%                   & 0.071                   & 0.099                    & 12.718\%                   & 0.071                   & 0.101                    & 12.675\%                    \\
                          & Step 10                   & 0.117                   & 0.157                    & 19.575\%                   & 0.129                   & 0.186                    & 21.522\%                   & 0.073                   & 0.103                    & 13.097\%                   & 0.071                   & 0.099                    & 12.653\%                    \\
                          & Step 11                   & 0.123                   & 0.165                    & 20.461\%                   & 0.129                   & 0.190                    & 21.556\%                   & 0.071                   & 0.100                    & 12.783\%                   & 0.070                   & 0.096                    & 12.562\%                    \\
                          & Step 12                   & 0.126                   & 0.174                    & 20.988\%                   & 0.127                   & 0.180                    & 21.168\%                   & 0.075                   & 0.105                    & 13.381\%                   & 0.073                   & 0.102                    & 13.181\%                    \\
                          & \textit{\textbf{Average}} & \textit{\textbf{0.113}} & \textit{\textbf{0.153}}  & \textit{\textbf{18.990\%}} & \textit{\textbf{0.118}} & \textit{\textbf{0.180}}  & \textit{\textbf{19.737\%}} & \textit{\textbf{0.071}} & \textit{\textbf{0.100}}  & \textit{\textbf{12.657\%}} & \textit{\textbf{0.072}} & \textit{\textbf{0.101}}  & \textit{\textbf{12.834\%}}  \\
\hline
\end{tabular}
\end{table*}

\begin{table*}
\centering
\caption{Comparative Analysis of LSTMSeq2Seq and LSTMSeq2SeqAtn Models in the Target Domain $T_{d2}$ for Forecasting Accuracy Over Different Steps in Two Scenarios: Before and After Data Augmentation. This table evaluates the performance metrics—Mean Absolute Error (MAE), Root Mean Square Error (RMSE), and Weighted Absolute Percentage Error (WAPE)—across forecast horizons of 6, 9, and 12 steps. It provides detailed values for each step and the average performance, highlighting the models' efficacy in predicting Target Domain $T_{d2}$.}
\label{table_target_domain_2}
\begin{tabular}{l|l|lll|lll||lll|lll} 
\hline
                          &                           & \multicolumn{6}{c||}{Target Domain $T_{d2}$ Before Data Augmentation}                                                                                             & \multicolumn{6}{c}{Target Domain $T_{d2}$ After Data Augmentation}                                                                                                 \\ 
\hline
                          &                           & \multicolumn{3}{c|}{LSTMSeq2Seq}                                                & \multicolumn{3}{c||}{LSTMSeq2SeqAtn}                                            & \multicolumn{3}{c|}{LSTMSeq2Seq}                                                & \multicolumn{3}{c}{LSTMSeq2SeqAtn}                                               \\ 
\hline
Forecast Horizon          &                           & \multicolumn{1}{c}{MAE} & \multicolumn{1}{c}{RMSE} & \multicolumn{1}{c|}{WAPE}  & \multicolumn{1}{c}{MAE} & \multicolumn{1}{c}{RMSE} & \multicolumn{1}{c||}{WAPE} & \multicolumn{1}{c}{MAE} & \multicolumn{1}{c}{RMSE} & \multicolumn{1}{c|}{WAPE}  & \multicolumn{1}{c}{MAE} & \multicolumn{1}{c}{RMSE} & \multicolumn{1}{c}{WAPE}    \\ 
\hline
\multirow{6}{*}{6 Step}   & Step 1                    & 0.199                   & 0.250                    & 24.675\%                   & 0.195                   & 0.250                    & 24.200\%                   & 0.125                   & 0.174                    & 17.001\%                   & 0.138                   & 0.190                    & 18.836\%                    \\
                          & Step 2                    & 0.181                   & 0.227                    & 22.665\%                   & 0.184                   & 0.232                    & 22.977\%                   & 0.117                   & 0.162                    & 15.932\%                   & 0.130                   & 0.177                    & 17.691\%                    \\
                          & Step 3                    & 0.184                   & 0.222                    & 23.019\%                   & 0.180                   & 0.219                    & 22.541\%                   & 0.115                   & 0.163                    & 15.696\%                   & 0.126                   & 0.174                    & 17.138\%                    \\
                          & Step 4                    & 0.182                   & 0.227                    & 22.936\%                   & 0.183                   & 0.230                    & 23.007\%                   & 0.113                   & 0.158                    & 15.425\%                   & 0.123                   & 0.170                    & 16.693\%                    \\
                          & Step 5                    & 0.178                   & 0.217                    & 22.616\%                   & 0.177                   & 0.219                    & 22.486\%                   & 0.112                   & 0.163                    & 15.227\%                   & 0.122                   & 0.170                    & 16.582\%                    \\
                          & Step 6                    & 0.177                   & 0.220                    & 22.377\%                   & 0.181                   & 0.225                    & 22.801\%                   & 0.117                   & 0.170                    & 15.859\%                   & 0.125                   & 0.176                    & 16.985\%                    \\
                          & \textit{\textbf{Average}} & \textit{\textbf{0.184}} & \textit{\textbf{0.227}}  & \textit{\textbf{23.048\%}} & \textit{\textbf{0.183}} & \textit{\textbf{0.229}}  & \textit{\textbf{23.002\%}} & \textit{\textbf{0.117}} & \textit{\textbf{0.165}}  & \textit{\textbf{15.857\%}} & \textit{\textbf{0.127}} & \textit{\textbf{0.176}}  & \textit{\textbf{17.321\%}}  \\ 
\hline
\multirow{9}{*}{9 Step}   & Step 1                    & 0.187                   & 0.236                    & 22.814\%                   & 0.174                   & 0.222                    & 21.253\%                   & 0.121                   & 0.167                    & 16.525\%                   & 0.132                   & 0.180                    & 17.974\%                    \\
                          & Step 2                    & 0.182                   & 0.226                    & 22.447\%                   & 0.173                   & 0.226                    & 21.318\%                   & 0.114                   & 0.154                    & 15.549\%                   & 0.122                   & 0.164                    & 16.630\%                    \\
                          & Step 3                    & 0.171                   & 0.210                    & 21.395\%                   & 0.166                   & 0.215                    & 20.675\%                   & 0.108                   & 0.149                    & 14.745\%                   & 0.116                   & 0.157                    & 15.781\%                    \\
                          & Step 4                    & 0.168                   & 0.211                    & 21.167\%                   & 0.164                   & 0.206                    & 20.731\%                   & 0.103                   & 0.141                    & 14.071\%                   & 0.114                   & 0.153                    & 15.474\%                    \\
                          & Step 5                    & 0.164                   & 0.209                    & 20.966\%                   & 0.169                   & 0.215                    & 21.549\%                   & 0.105                   & 0.145                    & 14.342\%                   & 0.112                   & 0.153                    & 15.257\%                    \\
                          & Step 6                    & 0.166                   & 0.210                    & 21.047\%                   & 0.175                   & 0.224                    & 22.256\%                   & 0.106                   & 0.144                    & 14.348\%                   & 0.114                   & 0.157                    & 15.562\%                    \\
                          & Step 7                    & 0.174                   & 0.219                    & 21.992\%                   & 0.182                   & 0.233                    & 22.960\%                   & 0.105                   & 0.144                    & 14.306\%                   & 0.112                   & 0.155                    & 15.191\%                    \\
                          & Step 8                    & 0.168                   & 0.217                    & 21.174\%                   & 0.176                   & 0.225                    & 22.090\%                   & 0.106                   & 0.147                    & 14.425\%                   & 0.115                   & 0.163                    & 15.576\%                    \\
                          & Step 9                    & 0.175                   & 0.222                    & 21.763\%                   & 0.185                   & 0.234                    & 23.059\%                   & 0.105                   & 0.149                    & 14.271\%                   & 0.115                   & 0.164                    & 15.614\%                    \\
                          & \textit{\textbf{Average}} & \textit{\textbf{0.173}} & \textit{\textbf{0.218}}  & \textit{\textbf{21.640\%}} & \textit{\textbf{0.174}} & \textit{\textbf{0.222}}  & \textit{\textbf{21.766\%}} & \textit{\textbf{0.105}} & \textit{\textbf{0.145}}  & \textit{\textbf{14.294\%}} & \textit{\textbf{0.117}} & \textit{\textbf{0.161}}  & \textit{\textbf{15.896\%}}  \\ 
\hline
\multirow{13}{*}{12 Step} & Step 1                    & 0.170                   & 0.219                    & 20.416\%                   & 0.157                   & 0.206                    & 18.870\%                   & 0.117                   & 0.160                    & 15.888\%                   & 0.135                   & 0.192                    & 18.410\%                    \\
                          & Step 2                    & 0.163                   & 0.219                    & 19.806\%                   & 0.166                   & 0.219                    & 20.182\%                   & 0.112                   & 0.153                    & 15.253\%                   & 0.130                   & 0.183                    & 17.675\%                    \\
                          & Step 3                    & 0.146                   & 0.204                    & 17.886\%                   & 0.160                   & 0.210                    & 19.633\%                   & 0.109                   & 0.150                    & 14.822\%                   & 0.123                   & 0.173                    & 16.686\%                    \\
                          & Step 4                    & 0.146                   & 0.195                    & 18.155\%                   & 0.174                   & 0.220                    & 21.545\%                   & 0.107                   & 0.149                    & 14.510\%                   & 0.122                   & 0.169                    & 16.551\%                    \\
                          & Step 5                    & 0.155                   & 0.202                    & 19.527\%                   & 0.186                   & 0.237                    & 23.458\%                   & 0.110                   & 0.152                    & 14.887\%                   & 0.122                   & 0.170                    & 16.593\%                    \\
                          & Step 6                    & 0.155                   & 0.201                    & 19.604\%                   & 0.173                   & 0.219                    & 21.921\%                   & 0.111                   & 0.156                    & 15.090\%                   & 0.119                   & 0.165                    & 16.196\%                    \\
                          & Step 7                    & 0.154                   & 0.203                    & 19.453\%                   & 0.183                   & 0.229                    & 23.187\%                   & 0.109                   & 0.155                    & 14.792\%                   & 0.116                   & 0.159                    & 15.742\%                    \\
                          & Step 8                    & 0.156                   & 0.201                    & 19.644\%                   & 0.179                   & 0.231                    & 22.624\%                   & 0.111                   & 0.161                    & 15.042\%                   & 0.116                   & 0.162                    & 15.736\%                    \\
                          & Step 9                    & 0.159                   & 0.204                    & 19.841\%                   & 0.185                   & 0.245                    & 23.094\%                   & 0.106                   & 0.152                    & 14.455\%                   & 0.112                   & 0.152                    & 15.189\%                    \\
                          & Step 10                   & 0.159                   & 0.205                    & 19.740\%                   & 0.188                   & 0.243                    & 23.273\%                   & 0.102                   & 0.143                    & 13.829\%                   & 0.110                   & 0.155                    & 14.952\%                    \\
                          & Step 11                   & 0.161                   & 0.205                    & 19.898\%                   & 0.184                   & 0.238                    & 22.699\%                   & 0.101                   & 0.141                    & 13.761\%                   & 0.113                   & 0.156                    & 15.291\%                    \\
                          & Step 12                   & 0.161                   & 0.207                    & 19.723\%                   & 0.198                   & 0.247                    & 24.149\%                   & 0.107                   & 0.147                    & 14.471\%                   & 0.111                   & 0.158                    & 15.048\%                    \\
                          & \textit{\textbf{Average}} & \textit{\textbf{0.157}} & \textit{\textbf{0.205}}  & \textit{\textbf{19.474\%}} & \textit{\textbf{0.178}} & \textit{\textbf{0.229}}  & \textit{\textbf{22.053\%}} & \textit{\textbf{0.106}} & \textit{\textbf{0.150}}  & \textit{\textbf{14.392\%}} & \textit{\textbf{0.119}} & \textit{\textbf{0.166}}  & \textit{\textbf{16.172\%}}  \\
\hline
\end{tabular}
\end{table*}

\begin{table*}
\centering
\caption{Comparative Analysis of LSTMSeq2Seq and LSTMSeq2SeqAtn Models in the Target Domain $T_{d3}$ for Forecasting Accuracy Over Different Steps in Two Scenarios: Before and After Data Augmentation. This table evaluates the performance metrics—Mean Absolute Error (MAE), Root Mean Square Error (RMSE), and Weighted Absolute Percentage Error (WAPE)—across forecast horizons of 6, 9, and 12 steps. It provides detailed values for each step and the average performance, highlighting the models' efficacy in predicting Target Domain $T_{d3}$.}
\label{table_target_domain_3}
\begin{tabular}{l|l|lll|lll||lll|lll} 
\hline
                                                         &         & \multicolumn{6}{c||}{Target Domain $T_{d3}$ Before Data Augmentation}                                                                                                    & \multicolumn{6}{c}{Target Domain $T_{d3}$ After Data Augmentation}                                                                                                       \\ 
\hline
                                                         &         & \multicolumn{3}{c|}{LSTMSeq2Seq}                                                & \multicolumn{3}{c||}{LSTMSeq2SeqAtn}                                            & \multicolumn{3}{c|}{LSTMSeq2Seq}                                                & \multicolumn{3}{c}{LSTMSeq2SeqAtn}                                               \\ 
\hline
\begin{tabular}[c]{@{}l@{}}Forcast\\Horizon\end{tabular} &         & \multicolumn{1}{c}{MAE} & \multicolumn{1}{c}{RMSE} & \multicolumn{1}{c|}{WAPE}  & \multicolumn{1}{c}{MAE} & \multicolumn{1}{c}{RMSE} & \multicolumn{1}{c||}{WAPE} & \multicolumn{1}{c}{MAE} & \multicolumn{1}{c}{RMSE} & \multicolumn{1}{c|}{WAPE}  & \multicolumn{1}{c}{MAE} & \multicolumn{1}{c}{RMSE} & \multicolumn{1}{c}{WAPE}    \\ 
\hline
\multirow{6}{*}{6 Step}                                  & Step 1  & 0.111                   & 0.141                    & 18.767\%                   & 0.116                   & 0.148                    & 19.600\%                   & 0.075                   & 0.108                    & 13.453\%                   & 0.072                   & 0.103                    & 12.823\%                    \\
                                                         & Step 2  & 0.106                   & 0.136                    & 18.078\%                   & 0.110                   & 0.139                    & 18.735\%                   & 0.071                   & 0.100                    & 12.765\%                   & 0.071                   & 0.099                    & 12.725\%                    \\
                                                         & Step 3  & 0.108                   & 0.142                    & 18.716\%                   & 0.109                   & 0.142                    & 18.839\%                   & 0.068                   & 0.094                    & 12.186\%                   & 0.072                   & 0.098                    & 12.811\%                    \\
                                                         & Step 4  & 0.109                   & 0.146                    & 18.917\%                   & 0.108                   & 0.147                    & 18.779\%                   & 0.067                   & 0.091                    & 11.966\%                   & 0.070                   & 0.097                    & 12.580\%                    \\
                                                         & Step 5  & 0.112                   & 0.150                    & 19.331\%                   & 0.106                   & 0.144                    & 18.274\%                   & 0.065                   & 0.090                    & 11.691\%                   & 0.069                   & 0.095                    & 12.380\%                    \\
                                                         & Step 6  & 0.105                   & 0.142                    & 18.043\%                   & 0.100                   & 0.133                    & 17.171\%                   & 0.068                   & 0.093                    & 12.175\%                   & 0.070                   & 0.096                    & 12.600\%                    \\
                                                         & \textit{\textbf{Average}} & \textit{\textbf{0.109}} & \textit{\textbf{0.143}}  & \textit{\textbf{18.642\%}} & \textit{\textbf{0.108}} & \textit{\textbf{0.142}}  & \textit{\textbf{18.566\%}} & \textit{\textbf{0.069}} & \textit{\textbf{0.096}}  & \textit{\textbf{12.373\%}} & \textit{\textbf{0.071}} & \textit{\textbf{0.098}}  & \textit{\textbf{12.653\%}}  \\ 
\hline
\multirow{9}{*}{9 Step}                                  & Step 1  & 0.105                   & 0.134                    & 17.440\%                   & 0.115                   & 0.153                    & 19.051\%                   & 0.073                   & 0.103                    & 12.952\%                   & 0.072                   & 0.102                    & 12.863\%                    \\
                                                         & Step 2  & 0.103                   & 0.131                    & 17.314\%                   & 0.125                   & 0.170                    & 21.139\%                   & 0.071                   & 0.098                    & 12.675\%                   & 0.072                   & 0.101                    & 12.915\%                    \\
                                                         & Step 3  & 0.104                   & 0.138                    & 17.731\%                   & 0.124                   & 0.180                    & 21.193\%                   & 0.069                   & 0.095                    & 12.277\%                   & 0.072                   & 0.097                    & 12.812\%                    \\
                                                         & Step 4  & 0.103                   & 0.138                    & 17.721\%                   & 0.116                   & 0.167                    & 19.936\%                   & 0.068                   & 0.094                    & 12.112\%                   & 0.069                   & 0.094                    & 12.397\%                    \\
                                                         & Step 5  & 0.106                   & 0.140                    & 18.209\%                   & 0.114                   & 0.156                    & 19.516\%                   & 0.067                   & 0.093                    & 12.064\%                   & 0.070                   & 0.097                    & 12.559\%                    \\
                                                         & Step 6  & 0.108                   & 0.141                    & 18.550\%                   & 0.117                   & 0.162                    & 20.113\%                   & 0.068                   & 0.095                    & 12.228\%                   & 0.072                   & 0.098                    & 12.807\%                    \\
                                                         & Step 7  & 0.107                   & 0.142                    & 18.361\%                   & 0.113                   & 0.158                    & 19.246\%                   & 0.069                   & 0.096                    & 12.392\%                   & 0.072                   & 0.101                    & 12.974\%                    \\
                                                         & Step 8  & 0.105                   & 0.140                    & 17.889\%                   & 0.109                   & 0.156                    & 18.442\%                   & 0.072                   & 0.103                    & 12.893\%                   & 0.076                   & 0.104                    & 13.554\%                    \\
                                                         & Step 9  & 0.107                   & 0.139                    & 18.086\%                   & 0.113                   & 0.160                    & 19.057\%                   & 0.073                   & 0.105                    & 13.064\%                   & 0.076                   & 0.106                    & 13.685\%                    \\
                                                         & \textit{\textbf{Average}} & \textit{\textbf{0.105}} & \textit{\textbf{0.138}}  & \textit{\textbf{17.922\%}} & \textit{\textbf{0.116}} & \textit{\textbf{0.163}}  & \textit{\textbf{19.744\%}} & \textit{\textbf{0.070}} & \textit{\textbf{0.098}}  & \textit{\textbf{12.459\%}} & \textit{\textbf{0.073}} & \textit{\textbf{0.100}}  & \textit{\textbf{12.996\%}}  \\ 
\hline
\multirow{13}{*}{12 Step}                                & Step 1  & 0.109                   & 0.136                    & 17.790\%                   & 0.102                   & 0.141                    & 16.626\%                   & 0.069                   & 0.098                    & 12.294\%                   & 0.075                   & 0.104                    & 13.358\%                    \\
                                                         & Step 2  & 0.107                   & 0.137                    & 17.714\%                   & 0.103                   & 0.153                    & 17.024\%                   & 0.071                   & 0.097                    & 12.576\%                   & 0.073                   & 0.102                    & 13.097\%                    \\
                                                         & Step 3  & 0.106                   & 0.142                    & 17.803\%                   & 0.113                   & 0.179                    & 18.863\%                   & 0.070                   & 0.096                    & 12.434\%                   & 0.076                   & 0.106                    & 13.628\%                    \\
                                                         & Step 4  & 0.105                   & 0.143                    & 17.636\%                   & 0.114                   & 0.179                    & 19.151\%                   & 0.069                   & 0.094                    & 12.251\%                   & 0.075                   & 0.105                    & 13.380\%                    \\
                                                         & Step 5  & 0.107                   & 0.146                    & 18.069\%                   & 0.116                   & 0.179                    & 19.646\%                   & 0.068                   & 0.094                    & 12.083\%                   & 0.075                   & 0.106                    & 13.421\%                    \\
                                                         & Step 6  & 0.107                   & 0.146                    & 18.095\%                   & 0.113                   & 0.178                    & 19.115\%                   & 0.070                   & 0.097                    & 12.540\%                   & 0.076                   & 0.108                    & 13.562\%                    \\
                                                         & Step 7  & 0.107                   & 0.145                    & 18.046\%                   & 0.110                   & 0.177                    & 18.708\%                   & 0.071                   & 0.100                    & 12.665\%                   & 0.076                   & 0.107                    & 13.515\%                    \\
                                                         & Step 8  & 0.105                   & 0.142                    & 17.704\%                   & 0.107                   & 0.163                    & 18.057\%                   & 0.072                   & 0.103                    & 12.822\%                   & 0.076                   & 0.109                    & 13.679\%                    \\
                                                         & Step 9  & 0.106                   & 0.143                    & 17.943\%                   & 0.107                   & 0.160                    & 18.059\%                   & 0.073                   & 0.106                    & 13.090\%                   & 0.078                   & 0.110                    & 13.906\%                    \\
                                                         & Step 10 & 0.109                   & 0.145                    & 18.295\%                   & 0.111                   & 0.161                    & 18.752\%                   & 0.076                   & 0.109                    & 13.526\%                   & 0.080                   & 0.112                    & 14.365\%                    \\
                                                         & Step 11 & 0.114                   & 0.153                    & 19.056\%                   & 0.114                   & 0.165                    & 19.031\%                   & 0.075                   & 0.109                    & 13.494\%                   & 0.080                   & 0.116                    & 14.422\%                    \\
                                                         & Step 12 & 0.119                   & 0.161                    & 19.828\%                   & 0.118                   & 0.164                    & 19.639\%                   & 0.076                   & 0.108                    & 13.552\%                   & 0.081                   & 0.115                    & 14.454\%                    \\
                                                         & \textit{\textbf{Average}} & \textit{\textbf{0.108}} & \textit{\textbf{0.145}}  & \textit{\textbf{18.165\%}} & \textit{\textbf{0.111}} & \textit{\textbf{0.167}}  & \textit{\textbf{18.556\%}} & \textit{\textbf{0.074}} & \textit{\textbf{0.106}}  & \textit{\textbf{13.191\%}} & \textit{\textbf{0.079}} & \textit{\textbf{0.111}}  & \textit{\textbf{14.057\%}}  \\
\hline
\end{tabular}
\end{table*}

\begin{figure*}[htp]
    \centering
    \begin{subfigure}{.32\textwidth}
        \centering
        \includegraphics[width=\linewidth]{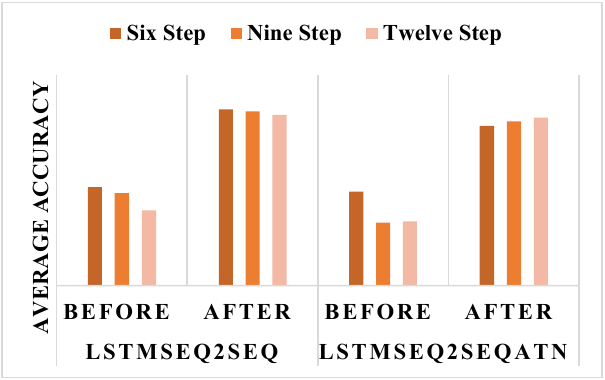}
        \caption{Target Domain $T_{d1}$}
        \label{fig:image1}
    \end{subfigure}%
    \hfill
    \begin{subfigure}{.32\textwidth}
        \centering
        \includegraphics[width=\linewidth]{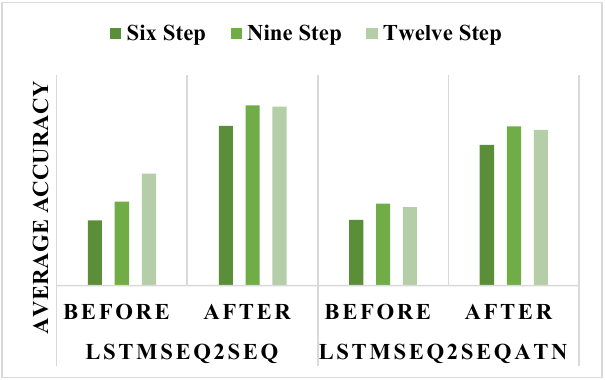}
        \caption{Target Domain $T_{d2}$}
        \label{fig:image2}
    \end{subfigure}%
    \hfill
    \begin{subfigure}{.32\textwidth}
        \centering
        \includegraphics[width=\linewidth]{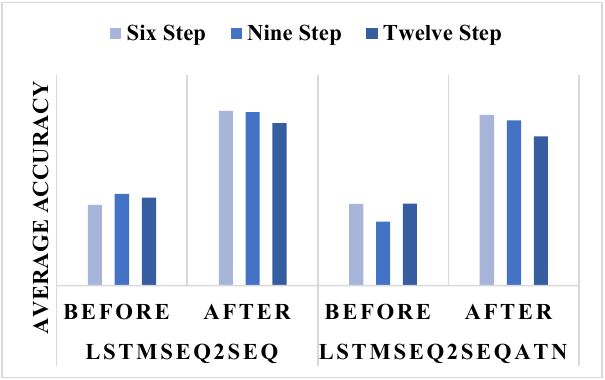}
        \caption{Target Domain $T_{d3}$}
        \label{fig:image3}
    \end{subfigure}
    \caption{Comparison of average accuracy between LSTMSeq2Seq and LSTMSeq2SeqAtn methods before and after data augmentation at six, nine, and twelve steps in the target domain $T_{d1}$, $T_{d2}$, and $T_{d3}$. The grouped bar chart demonstrates the improvement in prediction accuracy after data augmentation for both methods across different step intervals. The results indicate a consistent increase in accuracy after the application of the augmentation techniques.}
    \label{fig:average_acc}
\end{figure*}

\subsection{Comparative Analysis of Data Augmentation on Forecast Accuracy}

The Table \ref{tab:dtl_source_domain_performance} compares the average accuracy of two time series prediction models, LSTMSeq2Seq and LSTMSeq2SeqAttn, across three future steps in source domain. The data indicates a common trend of diminishing accuracy as predictions extend further into the future for both models, with the LSTMSeq2Seq model exhibiting a slight but consistent decline from 93.72\% to 91.20\% from Step 6 to Step 12. The LSTMSeq2SeqAttn model, which incorporates an attention mechanism, demonstrates marginally better accuracy at Steps 9 and 12, suggesting that the attention mechanism may provide more context for predictions over longer horizons. However, the differences between the models are minimal, with the largest observed being only 0.06\% at Step 12. Now, we perform a comparative analysis examining the impact of data augmentation on the forecasting accuracy of LSTMSeq2Seq and LSTMSeq2SeqAttn models within three distinct target domains, \(T_{d1}\), \(T_{d2}\), and \(T_{d3}\). Each domain's performance is measured by the Mean Absolute Error (MAE), Root Mean Square Error (RMSE), and Weighted Absolute Percentage Error (WAPE) across forecast horizons of 6, 9, and 12 steps.

For the 6-step forecast in target domain $T_{d1}$ as shown in Table \ref{table_target_domain_1}, the LSTMSeq2Seq model demonstrated a notable reduction in MAE from 0.102 to 0.069, indicating a 32.35\% improvement post data augmentation. The RMSE and WAPE followed suit, decreasing from 0.134 to 0.099 (26.12\% improvement) and from 16.95\% to 12.26\% (27.66\% improvement), respectively. The LSTMSeq2SeqAttn model showed similar trends, with the MAE decreasing from 0.103 to 0.074, RMSE from 0.134 to 0.101, and WAPE from 17.75\% to 13.29\%. The enhancements are consistent across the 9 and 12-step forecasts as well, although the magnitude of improvement diminishes with the increase in forecast length, which is a common phenomenon due to the inherent increase in uncertainty in longer-term predictions. In target domain $T_{d2}$, the LSTMSeq2Seq model's MAE decreased from 0.184 to 0.125 (32.07\% improvement) in the 6-step forecast after data augmentation, whereas LSTMSeq2SeqAttn's MAE dropped from 0.183 to 0.138 (24.59\% improvement) as shown in Table \ref{table_target_domain_2}. In this domain, the LSTMSeq2Seq model also displayed a greater reduction in RMSE and WAPE compared to LSTMSeq2SeqAttn, indicating a more significant benefit from the augmented data. The average MAE for LSTMSeq2Seq in the 6-step forecast decreased from 0.109 to 0.069 (36.70\% improvement), with the RMSE and WAPE reflecting similar trends in target domain $T_{d3}$. The LSTMSeq2SeqAttn model exhibited a smaller decrease in MAE from 0.108 to 0.071 (34.26\% improvement) as indicated in Table \ref{table_target_domain_3}. Interestingly, in this domain, the attention mechanism seems to offer less incremental benefit post data augmentation compared to domains $T_{d1}$ and $T_{d2}$.

Across all domains, it is evident in Figure \ref{fig:average_acc} that data augmentation considerably improves the forecasting accuracy of both models, with the LSTMSeq2Seq model generally showing a slightly higher percentage decrease in all error metrics. This could suggest that the LSTMSeq2Seq model has a greater capacity to leverage additional data to improve its predictive accuracy. It is also noteworthy that the benefits of data augmentation are more pronounced in the shorter-term forecasts, with diminishing returns as the forecast horizon extends. This observation aligns with theoretical expectations, as longer forecast periods inherently incorporate more uncertainty, thereby potentially diluting the impact of augmented data. Furthermore, while the LSTMSeq2SeqAttn model benefits from the attention mechanism's ability to focus on more salient features, the incremental benefit over the LSTMSeq2Seq model is less pronounced after data augmentation, which could imply that the attention mechanism's effectiveness is somewhat contingent on the data's complexity and variety.

In summary, our results underscore the efficacy of data augmentation in enhancing the predictive accuracy of LSTM-based forecasting models, suggesting that data augmentation should be a key consideration in the development of robust forecasting systems.

\begin{figure*}[!htbp]
\centering
    \includegraphics[width=16cm,height = 5cm]{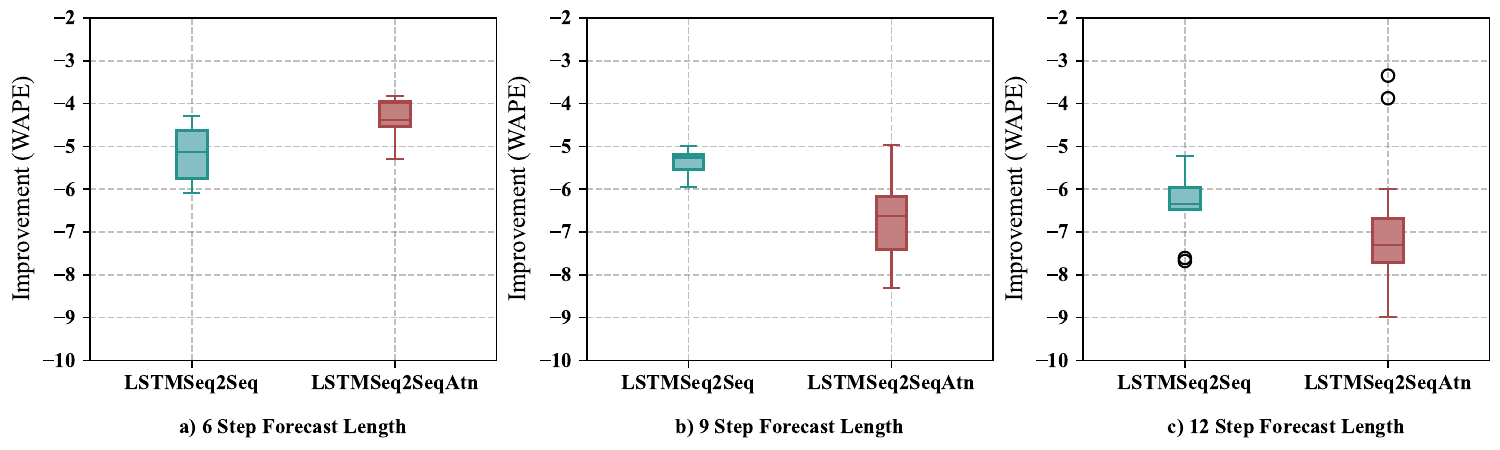}
    \caption{Boxplots showing the improvement in forecast accuracy in the target domain $T_{d1}$ over three forecast lengths (6 Step, 9 Step, and 12 Step) for two models: LSTMSeq2Seq and LSTMSeq2SeqAtn. The improvement is measured in terms of WAPE. Each boxplot represents the distribution of improvement values, with median lines marked in green for LSTMSeq2Seq and in red for LSTMSeq2SeqAtn.}
   \label{fig:box_plot_target_domain_1}
\end{figure*}

\begin{figure*}[!htbp]
\centering
    \includegraphics[width=16cm,height = 5cm]{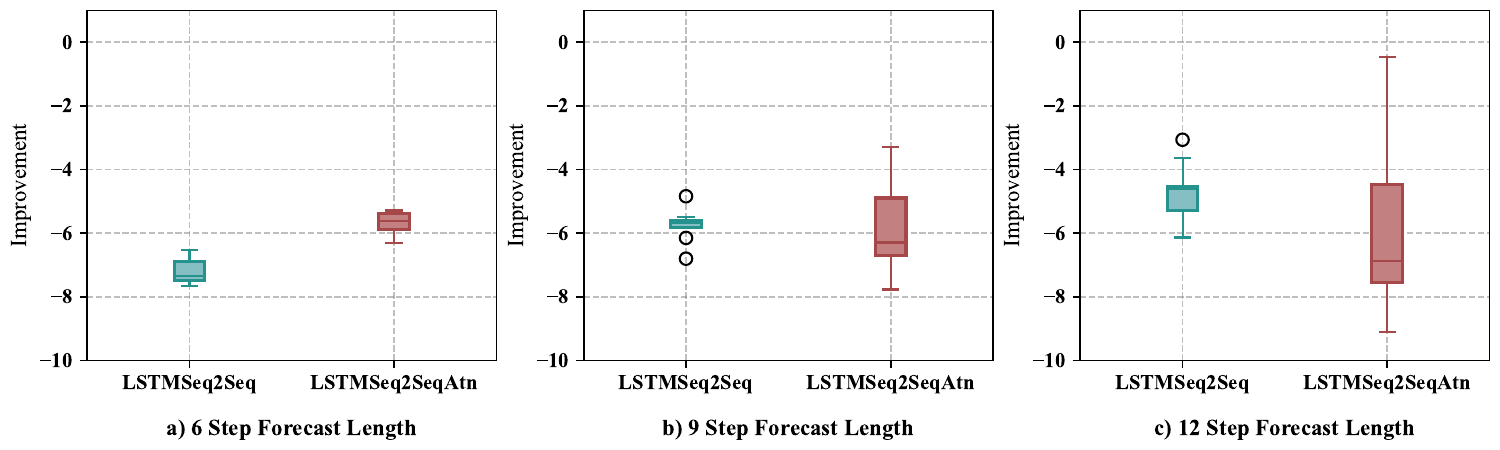}
    \caption{Boxplots showing the improvement in forecast accuracy in the target domain $T_{d2}$ over three forecast lengths (6 Step, 9 Step, and 12 Step) for two models: LSTMSeq2Seq and LSTMSeq2SeqAtn. The improvement is measured in terms of WAPE. Each boxplot represents the distribution of improvement values, with median lines marked in green for LSTMSeq2Seq and in red for LSTMSeq2SeqAtn.}
     \label{fig:box_plot_target_domain_2}
\end{figure*}

\begin{figure*}[!htbp]
\centering
    \includegraphics[width=16cm,height = 5cm]{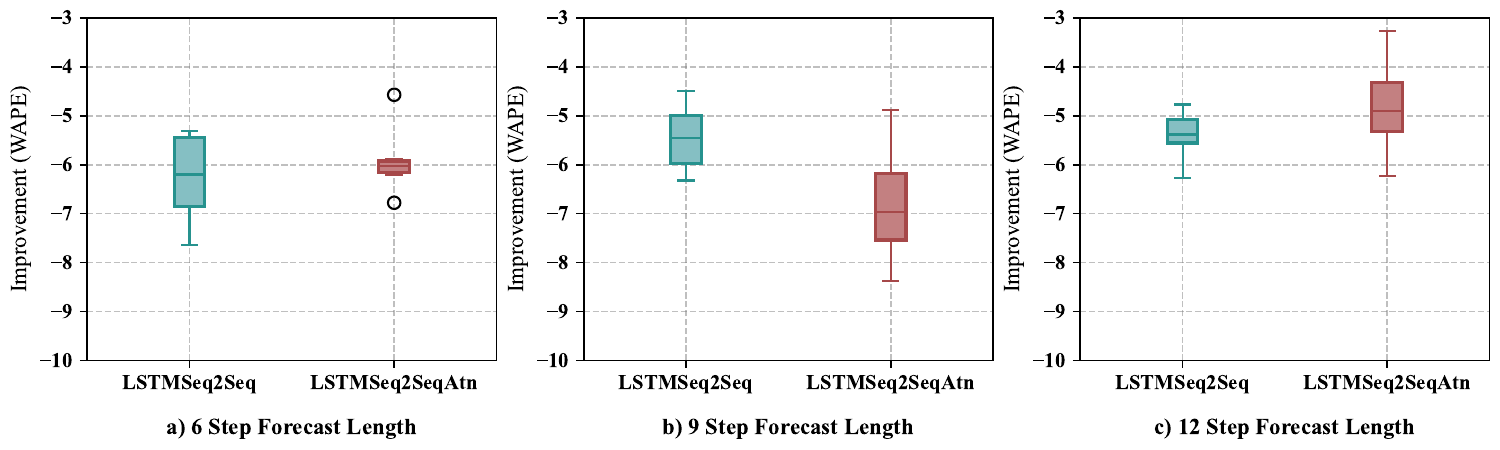}
    \caption{Boxplots showing the improvement in forecast accuracy in the target domain $T_{d3}$ over three forecast lengths (6 Step, 9 Step, and 12 Step) for two models: LSTMSeq2Seq and LSTMSeq2SeqAtn. The improvement is measured in terms of WAPE. Each boxplot represents the distribution of improvement values, with median lines marked in green for LSTMSeq2Seq and in red for LSTMSeq2SeqAtn.}
    \label{fig:box_plot_target_domain_3}
\end{figure*}

\subsection{Analysis of Variability and Consistency of Prediction Model}
The Interquartile Range (IQR) is a pivotal statistic for evaluating the consistency of model improvements \cite{goetz2015evaluating}. For the LSTMSeq2Seq and LSTMSeq2SeqAtn models, the IQR provides insight into the central 50\% spread of performance gains across different step forecasts. We have considered the variability in model performance by presenting the IQR of the cross-domain outcomes. Reduced IQR values signified greater consistency in model performances.

In the analysis of model performance across various forecasting horizons in target domain $T_{d1}$, the Interquartile Range (IQR) serves as a robust measure of the variability inherent in the improvements rendered by the models depicted in Figure \ref{fig:box_plot_target_domain_1}. For the 6-step forecast, the LSTMSeq2Seq model exhibits an IQR of approximately 1.1305, indicating that the middle 50\% of its performance enhancements are distributed within this range. In contrast, the inclusion of an attention mechanism in the LSTMSeq2SeqAtn model results in a reduced IQR of about 0.578, signifying a more concentrated clustering of enhancement values and, by implication, a more consistent performance. Moving to the 9-step forecast, the LSTMSeq2Seq model's performance consistency is particularly higher, with an IQR of just 0.349, the narrowest among the configurations studied. Conversely, the LSTMSeq2SeqAtn model's IQR expands to 1.237, denoting a wider dispersion of performance improvements. The 12-step forecasts follow a similar pattern; the LSTMSeq2Seq model demonstrates moderate variability with an IQR of approximately 0.5298, whereas the LSTMSeq2SeqAtn model's IQR increases to 1.0238, highlighting a less consistent performance. These findings highlight the varying impacts of attention mechanisms on model reliability across different forecast lengths, with the attention layer integrated seq2seq models displaying greater variability in their performance improvements.

For the 6-step forecast horizon in target $T_{d2}$ shown in Figure \ref{fig:box_plot_target_domain_2}, the LSTMSeq2Seq model manifests an IQR of 0.6, suggesting moderate dispersion in the central 50\% of its performance improvement distribution. The LSTMSeq2SeqAtn model yields a slightly narrower IQR of approximately 0.508, indicating a marginally more consistent performance within its median range. The discrepancy narrows further at the 9-step forecast, where the LSTMSeq2Seq model achieves an IQR of 0.219, reflective of a stronger cluster of improvement values. In contrast, the LSTMSeq2SeqAtn model for the same step length exhibits an IQR of 1.8, a substantial increase that indicates a far broader spread in performance improvements. This trend of increasing variability with attention mechanisms increases at the 12-step forecast, where the LSTMSeq2Seq model records an IQR of 0.761, while LSTMSeq2SeqAtn, displays a significantly higher IQR of 3.07775. In the comparative analysis of LSTMSeq2Seq models over varying forecast horizons for target $T_{d2}$, it was found that the attention mechanism's impact on performance consistency varied with forecast length. At a 6-step forecast, both models demonstrated moderate variability, with the LSTMSeq2SeqAtn model showing slightly tighter performance clustering. However, at a 9-step forecast, the LSTMSeq2Seq model exhibited a marked increase in consistency, while the LSTMSeq2SeqAtn's performance variability significantly widened. This divergence was even more evident at the 12-step forecast, where the LSTMSeq2SeqAtn's variability more than quadrupled compared to the 6-step forecast, highlighting a trend where the attention mechanism may introduce greater variability as the forecasting interval lengthens.

At a 6-step forecast in target domain $T_{d3}$ illustrated in Figure \ref{fig:box_plot_target_domain_3}, the LSTMSeq2Seq model exhibits a relatively high IQR of 1.39325, suggesting a considerable spread in its performance improvements, indicative of variability in its predictive consistency. In contrast, its attention-enhanced variant, the LSTMSeq2SeqAtn model, achieves a substantially lower IQR of 0.23325 at the same forecast horizon, pointing towards a more stable and consistent improvement in its predictions. Extending the forecast to 9 steps, the LSTMSeq2Seq model shows an IQR of 0.973, a decrease from the 6-step forecast but still reflective of significant variability. The LSTMSeq2SeqAtn model for the same step length registers an IQR of 1.351, which, while higher than its 6-step counterpart, implies a spread in performance that could be attributed to the model capturing more complex data patterns or to over-fitting. At the 12-step forecast, the LSTMSeq2Seq model's IQR narrows further to 0.48275, suggesting a convergence in its performance improvements and possibly a better fit for longer-term predictions. The LSTMSeq2SeqAtn model, on the other hand, presents an IQR of 0.99275, which is lower than the 9-step but higher than the 6-step forecast, indicating a non-linear relationship between the forecast horizon and performance consistency when the attention mechanism is applied. In comparing LSTMSeq2Seq models with and without attention mechanisms across different forecast horizons, the attention LSTMSeq2SeqAtn initially demonstrate greater consistency with lower IQR values for short-term forecasts. However, as the forecast horizon extends, these models exhibit increased variability, often exceeding that of the standard LSTMSeq2Seq models. This trend suggests that while attention mechanisms can enhance short-term forecasting by focusing on critical data points, their performance advantage may wane over longer sequences, potentially due to over-fitting or misalignment with the temporal structure of the data. 

In summary, the extensive analysis across different forecasting horizons within target domains $T_{d1}$, $T_{d2}$, and $T_{d3}$ has explained a potential pattern regarding the application of attention mechanisms in LSTMSeq2Seq models. It is evident that attention mechanisms tend to enhance the consistency of performance improvements in the short-term (6-step forecasts), as indicated by lower IQR values across the domains. This suggests that the attention mechanism's ability to focus on significant parts of the input sequence yields more stable outcomes when the forecasting horizon is relatively narrow. However, as the forecasting horizon expands (9-step and 12-step forecasts), the performance advantage presented by attention mechanisms appears to diminish, with the models often demonstrating increased variability as reflected by higher IQRs. In particular, at the 12-step forecast, the performance of the LSTMSeq2SeqAtn models shows a considerable spread, indicating less consistency and potential over-fitting or challenges in managing the complexity inherent in longer sequences.

\subsection{Analysis of Model Outliers}
The outlier analysis across three domains for different LSTM models over various forecast lengths provides a comparative perspective on model performance as measured by the improvement in Weighted Absolute Percentage Error (WAPE).

In domain $T_{d1}$, the 6-step and 9-step forecast lengths exhibited no outliers for either the LSTMSeq2Seq or LSTMSeq2SeqAtn models, indicating consistent model performance within the expected range. However, the 12-step forecast length revealed outliers for both models. Specifically, the LSTMSeq2Seq model had an outlier that showed a significant improvement in accuracy, while the LSTMSeq2SeqAtn model had two outliers indicating less improvement than average. Domain $T_{d2}$'s analysis aligns with domain $T_{d1}$ for the 6-step forecast length, with no outliers reported. The 9-step forecast for the LSTMSeq2Seq model diverged with three outliers that outperformed the general trend, suggesting exceptionally accurate forecasts. The 12-step forecast mirrored the 9-step for the LSTMSeq2Seq model with one outlier, again reflecting a better-than-average performance. The LSTMSeq2SeqAtn model showed no outliers for the 9-step and 12-step forecasts. In domain $T_{d3}$, the 6-step forecast length presented a contrast, with the LSTMSeq2SeqAtn model displaying two outliers, both indicating subpar performance improvements. For the 9-step forecast length, the LSTMSeq2Seq model maintained a consistent performance with no outliers, while the LSTMSeq2SeqAtn model showed one outlier with a lower-than-typical improvement. Both models had no outliers in the 12-step forecast length.

Summarizing these findings, the LSTMSeq2Seq model generally showed a trend of occasional exceptionally high accuracy in longer forecast lengths, while the LSTMSeq2SeqAtn model's outliers were indicative of under-performance in the shorter forecast horizon. These variations in performance improvement suggest that while LSTMSeq2Seq might be capturing certain data patterns more effectively in longer forecasts, the LSTMSeq2SeqAtn may require adjustments or further investigation into the causes of its outliers, particularly in the 6-step forecast domain.

\section{Conclusion}
\label{conclusion}
In this research, we utilized transfer learning techniques and data augmentation methods to predict internet traffic patterns in smaller network datasets. Our investigation aimed to alleviate the challenges of scarce data in smaller Internet Service Provider (ISP) networks, which makes the development of reliable prediction models a difficult task. The models LSTMSeq2Seq and LSTMSeq2SeqAtn, originally trained on larger datasets, were transferred and fine-tuned on smaller datasets. This approach successfully reduced the need for extensive data collection in smaller ISP networks, saving time and resources.

The findings revealed that both LSTMSeq2Seq and LSTMSeq2SeqAtn models perform comparably well on single-step predictions in the larger dataset, indicating their robustness and reliability. In contrast, the models struggled with multi-step predictions, revealing inherent challenges in long-term forecasts due to accumulated errors over time. Upon transfer to the smaller datasets, it was observed that LSTMSeq2Seq performed generally better than LSTMSeq2SeqAtn. The attention mechanism in LSTMSeq2SeqAtn did not seem to provide additional benefits for the prediction tasks at hand, suggesting that increasing model complexity does not always lead to improved performance. It is crucial to consider the specific characteristics and requirements of the task when selecting a model. The significant difference in dataset sizes between the source domain and target domains may have affected the models' ability to generalize to the smaller target domains. The models' performance varied across different target domains, suggesting that each domain has distinct characteristics that influence model performance.

Data augmentation via the Discrete Wavelet Transform (DWT) was utilized to enhance the target dataset size. This significantly improved the performance of both models, suggesting that data augmentation techniques can play a crucial role in improving model performance, especially when dealing with limited data. Despite the promising results, the research has limitations that provide directions for future work. First, the attention mechanism in the LSTMSeq2SeqAtn model may need further fine-tuning or alternative approaches to improve its performance. Second, while DWT was effective in this study, other data augmentation techniques could also be explored to further increase the diversity and representativeness of the smaller datasets. Third, we might also need to look into further improving the multi-step prediction performance of these models, as it is crucial for longer-term projections. Lastly, the discrepancies observed in the performance across different target domains suggest a need for further investigation into domain-specific features. Understanding these features better could help design models that are more adaptable and thus more accurate for each specific domain. We also plan to explore other architectures and strategies to handle the uncertainty associated with longer prediction horizons. 


\bibliographystyle{IEEEtran}
\bibliography{ref}

\newpage

\section{Biography Section}
\begin{IEEEbiography}[{\includegraphics[width=1in,height=1.25in,clip,keepaspectratio]{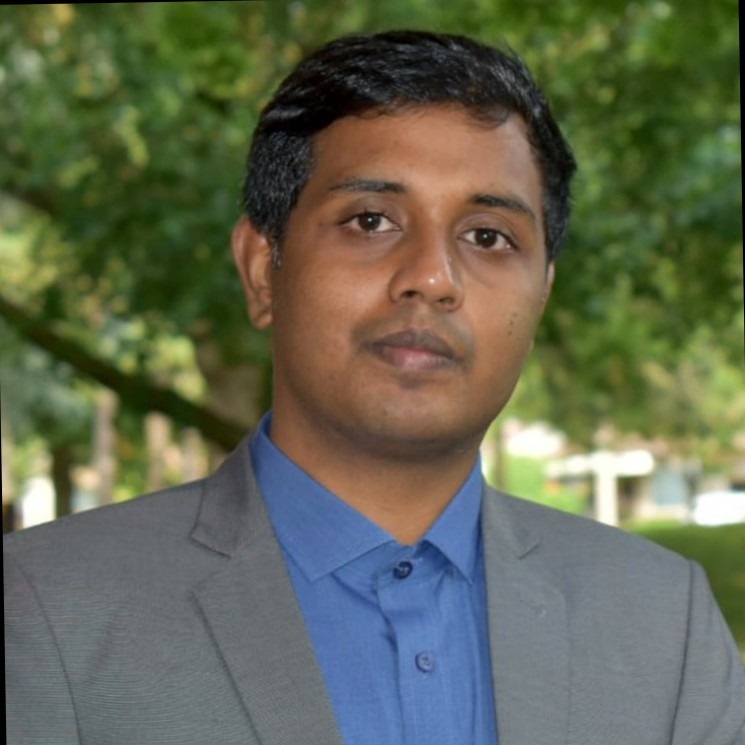}}]{Sajal Saha, Ph.D.} is an Assistant Professor in the Department of Computer Science at the University of Northern British Columbia, Canada. His research interests are primarily focused on Internet Traffic Analysis, Cyber-attack Detection, and Data Mining, with special emphasis on traffic prediction, anomaly detection, and association rule mining. He completed his Ph.D. at Western University, Canada, where his research contributed significantly to efficient traffic forecasting and cyber-attack detection. Alongside his academic contributions, he has also gained considerable industry experience as a Research Fellow at Juniper Network and Software Engineer at Samsung Electronics. 
\end{IEEEbiography}

\begin{IEEEbiography}[{\includegraphics[width=1in,height=1.25in,clip,keepaspectratio]{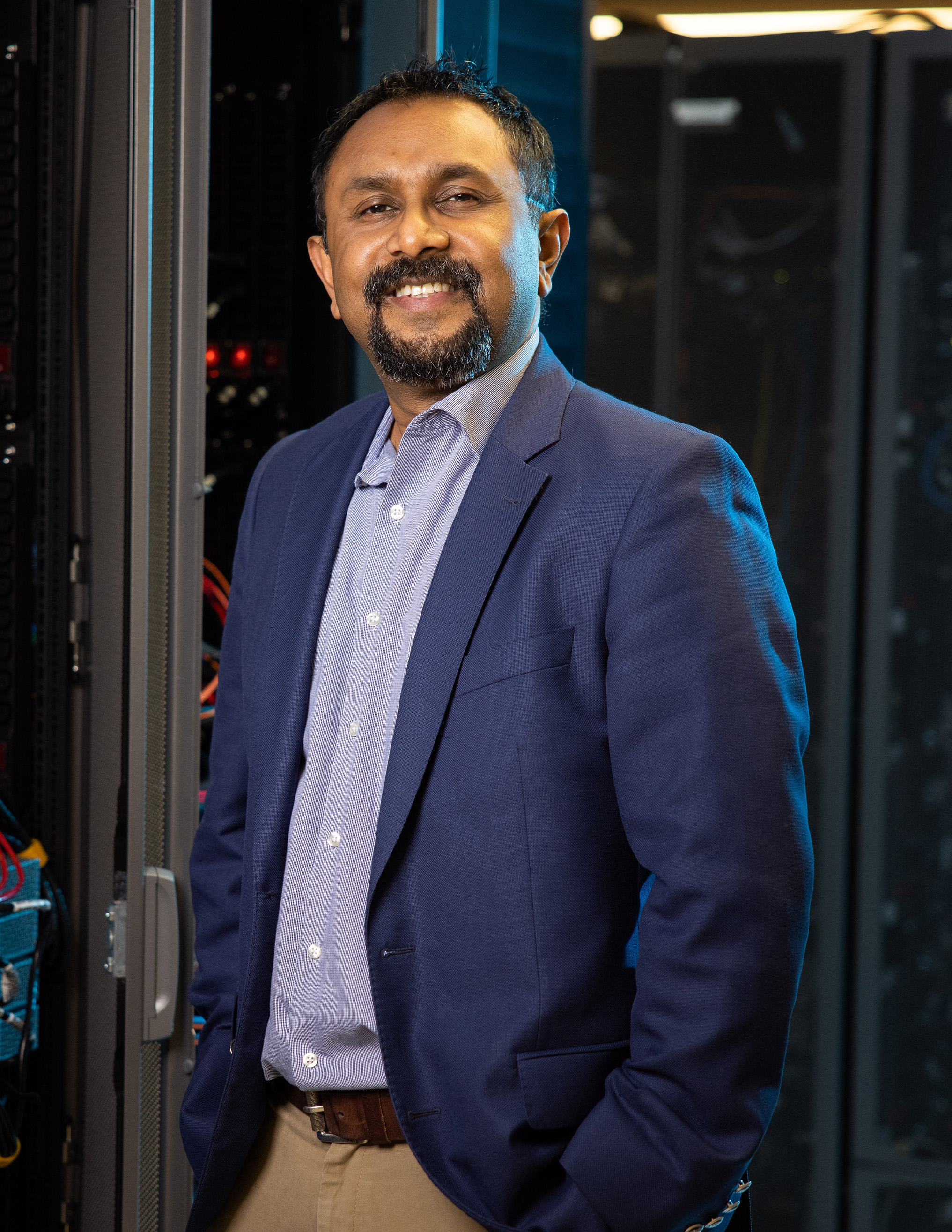}}]{Dr. Anwar Haque} is a faculty member in the Dept. of Computer Science at the University of Western Ontario, Canada. Before joining Western, he was an Associate Director at Bell Canada. He is a leading international expert on next-generation communication network resources and performance management, cyber security, and smart city applications. Dr. Haque has authored/co-authored over 80 peer-reviewed research publications in leading journals and conferences, authored many industry technical papers, and held a number of patents/licenses. He has been awarded several national/provincial-level research grants, including NSERC, MITACS, OCE, and SOSCIP. Dr. Haque’s collaborative research grants are valued at more than \$15 million. Dr. Haque is serving on the inaugural advisory committee for the newly established Bell-Western 5G research centre, and he established an industry consortium to promote and support smart systems and digital services research at Western. Dr. Haque is the director of the Western Information and Networking Group (WING) Lab at Western.
\end{IEEEbiography}

\begin{IEEEbiography}[{\includegraphics[width=1in,height=1.25in,clip,keepaspectratio]{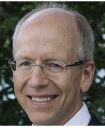}}]{Greg Sidebottom} received his PhD. degree in Computer
Science from Simon Fraser University. He is a Principal Engineer at Juniper Networks. He is the architect of Juniper’s industry pioneering Software Defined Network and Network Function Virtualization products. Greg is currently advancing the state of the art in segment routing and peer engineering with Juniper’s Northstar WAN SDN controller.
\end{IEEEbiography}

\end{document}